\documentclass{article}
\usepackage{amssymb, amsmath, latexsym}
\newtheorem{theorem}{Theorem}
\newtheorem{lemma}{Lemma}
\newtheorem{proposition}{Proposition}

\newtheorem{definition}{Definition}
\newtheorem{notation}{Notation}
\newtheorem{remark}{Remark}
\begin{document}
\title{\bf Period-halving Bifurcation of a Neuronal Recurrence Equation}
\author{Ren\'{e} Ndoundam${}^{a,b}$  \\
${}^{a}${\small University  of Yaounde I, UMI 209, UMMISCO, P.o.Box  337 Yaounde, }  \\
${}^{}${\small    Cameroon  }  \\
${}^{b}${\small University  of Yaounde I, LIRIMA,   Team  GRIMCAPE,  Faculty of Science,  } \\
${}^{}${\small   Department  of  Computer  Science, P.o. Box  812 Yaounde, Cameroon } \\
{\small E.mail : ndoundam@gmail.com , ndoundam@yahoo.com } 
             }
\date{}
\maketitle { }
\begin{abstract}  We study the sequences generated by neuronal recurrence 
equations of the form $x(n) = {\bf 1}[ \sum_{j=1}^{h} a_{j} x(n-j)- \theta ]$. From a neuronal recurrence 
equation of memory size $h$ which describes a cycle of length $\rho(m) \times lcm(p_0, p_1 , \dots, p_{-1+\rho(m)})$, we construct
 a set of $\rho(m)$ neuronal recurrence equations whose dynamics describe respectively the transient of length
 $O( \rho(m) \times  lcm(p_0, \dots, p_{d}))$ and the cycle of length $O(  \rho(m) \times   lcm(p_{d+1},  \dots, p_{-1+\rho(m)}))$ if $0 \leq d \leq -2+\rho(m)$ and 1 if $d=\rho(m)-1$. 
 This result shows  the exponential time of the convergence  of neuronal recurrence equation to fixed points  and the existence of the period-halving bifurcation.
\end{abstract}
\section{Introduction}
Caianiello and De Luca \cite{CL 66} have suggested that the
dynamic behavior of a single neuron with a memory, which does not
interact with other neurons can be modeled by the following
recurrence equation :
\begin{gather}   \label{eeq:n1}
x(n) = {\bf 1} \left[ \sum_{j=1}^{k}a_{j}x(n-j)- \theta \right]
\end{gather}
where :
\begin{itemize} \item[$\bullet$] $x(n)$ is a variable representing
the state of the neuron at $t \ = \ n$.
\end{itemize}
\begin{itemize} \item[$\bullet$] $x(0), x(1), \cdots, x(k-2), x(k-1)$ are
the initial states.
\end{itemize}
\begin{itemize}
\item[$\bullet$] $k$ is the memory length, i.e., the state of the
neuron at time $t \ = \ n$ depends on the states $x(n-1), \dots ,
x(n-k)$ assumed by the neuron at the $k$ previous steps $t \ = \
n-1, \dots, n-k$.
\end{itemize}
\begin{itemize}
\item[$\bullet$] $a_j$ ($j = 1, \dots ,k$) are real numbers called the
weighting coefficients. More precisely, $a_j$ represents the
influence of the state of the neuron at time $n-j$ on the state
assumed by the neuron at time $n$.
\end{itemize}
\begin{itemize} \item[$\bullet$] $\theta$ is a real number called the
threshold. \end{itemize}
\begin{itemize} \item[$\bullet$]
{\bf 1}[$u$] = 0 if $u \ < \ 0$, and {\bf 1}[$u$] = 1 if  $u \geq 0$.
\end{itemize}
The system obtained by interconnecting several neurons is called
a neural network. These networks were introduced by McCulloch
and Pitts \cite{MP 43}, and are quite powerful. Neural networks
are able to simulate any sequential machine  or Turing machine if
an infinite number of cells is provided. Neural networks
have been studied extensively as tools for solving various
problems such as classification, speech recognition, and image
processing \cite{Fogl 87}. The field of application of threshold
functions is large\cite{GM 90, Ho 82, HoT 85, Fogl 87} . The spin
moment of the spin glass system is one of the most cited example in
solid state physics that has been simulated by neural networks. \\
Neural networks are usually implemented by using electronic components or
is simulated in software on a digital computer. One way in which the collective
properties of a neural network may be used to implement a computational task is 
by way of the concept of {\it energy minimization}. The Hopfield network is a well-known 
example of such an approach. It has attracted  great  attention 
in  literature as a {\it content-addressable memory} \cite{SH 99}.  \\
Given a finite neural network, the configuration assumed by the
system at time t is ultimately periodic. As a consequence, there
is an integer $p \ > \ 0$ called the period (or a length of a
cycle) and another integer $T \geq 0$ called the transient length such that:
\begin{list}{\texttt{$\bullet$}}{}
\item $Y(p+T) = Y(T)$
\item $\nexists \ T' \ and \ p' \ \ (T', p') \ne (T, p) \ \ \ T \geq T' \ and
        \ p \geq p' \ \ {\rm such \ that} \ Y(p'+T') \ = \ Y(T')$
\end{list}
where $Y(t) = (x(t), x(t-1), \dots , x(t-k+2), x(t-k+1))$. The
period and the transient length of the sequences generated are good
measures of the complexity of the neuron. A bifurcation occurs when a small smooth change made to the 
parameter values (the bifurcation parameters) of a system causes a sudden 'qualitative' or topological
 change in its behaviour. A period halving bifurcation in a dynamical system is a bifurcation 
 in which the system switches to a new behaviour with half the period of the original system. A great variety of
 results have been established on recurrence equations modeling neurons with memory \cite{GM 90, CMG 88, CTT 92, 
CG 84, NM 00a, NM 00b, NT 04, TT 93, Mou 89}. However  some mathematical properties are still very intriguing and many problems are being posed. 
 For example, the question remains as to whether there exists one neuronal recurrence equation with transients of exponential lengths \cite{Cos 99}. In \cite{NT 03}, we give a positive answer to this question by exhibiting a neuronal recurrence equation with
 memory which generates a sequence of exponential transient length and exponential period length with respect to the memory length.
Despite this positive answer, one question remains:  does  there exist one neuronal recurrence equation with exponential transient length and fixed point ?  \\
In this work, from a neuronal recurrence equation of memory size $(6m-1) \times (\rho(m))^2$, whose dynamics contains a cycle of length  $ \rho(m) \times lcm(p_0, \ p_1, \ \dots \ , p_{-1+\rho(m)})$, we build a set of $\rho(m)$ neuronal recurrence equations whose dynamics describe respectively:
\begin{itemize}
\item the transient of length $\rho(m) \times lcm(p_0, p_1, \dots, p_{d})+h+d+1-\Bigl( \rho(m) \times (1+p_d) \Bigl)$,  if  $0 \leq d \leq -1+\rho(m)$
\item the cycle of length $\rho(m) \times lcm(p_{d+1}, p_{d+2}, \dots, p_{-1+\rho(m)})$ if $0 \leq d \leq -2+\rho(m)$ and 1 if $d=\rho(m)-1$
\end{itemize}
Thus, we give a positive answer to the precedent question. \\
The technique used in this paper to get  the period-halving bifurcation is to modify some parameters  (weighting coefficients and threshold) of the neuronal recurrence
equation. This technique relies on control theory. Controllability is related to the possibility of forcing the system into a particular state by using an 
 appropriate control signal.   \\

The paper is organized as follows: in Section 2, some previous 
results are presented. Section 3 presents some preliminaries. Section 4 is devoted to the construction of neuronal 
recurrence equation $z(n,d)$. Section 5 deals with the behavior of neuronal recurrence equation $z(n,d)$. Concluding remarks are stated in Section 6.
\section{Previous  Results}

The only study of bifurcation was done by Cosnard and Goles in \cite{CG 84}. Cosnard and Goles
 \cite{CG 84} studied the bifurcation in two particular cases of neuronal recurrence equation: \\
{\bf Case 1:}  geometric coefficients and bounded memory  \\
Cosnard and Goles described completely the structure of the bifurcation of the following equation:
\[
x_{n+1} = {\bf 1} \left[  \theta - \sum_{i=0}^{k-1} b^i x_{n-i} \right]
\]
when $\theta$ varies. They showed that the associated rotation number is an increasing number of the parameter $\theta$. \\
{\bf Case 2:}  geometric coefficients and unbounded memory  \\
Cosnard and Goles described completely the structure of the bifurcation of the following equation:
\[
x_{n+1} = {\bf 1} \left[  \theta - \sum_{i=0}^{n} b^i x_{n-i} \right]
\]
when $\theta$ varies. They showed that the associated rotation number is a devil's staircase.  \\

From line 11 to line 15 of page 15 in \cite{CTT 92}, it is written: ``This shows that, if there
is a neuronal recurrence equation with memory length $k$ that generates sequences 
of periods $p_1, \dots, p_{r}$, then there is a neuronal recurrence equation 
with memory length $kr$ that generates a sequence of period 
$lcm(p_1, \dots , p_r )r$, where $lcm$ denotes the least common multiple.''
This allows us to write the following fundamental lemma of composition of
a neuronal recurrence equation:
\begin{lemma} \cite{CTT 92}   \label{elll:nn1} \\
If there is a neuronal recurrence equation with memory length $k$ that generates 
sequences of periods $p_1, p_2,  \dots , p_r$, then there is a neuronal recurrence
equation with memory length $kr$ that generates a sequence of period
$r \cdot lcm(p_1, \cdots , p_r)$.
\end{lemma}    
$\blacksquare$  \\
Lemma \ref{elll:nn1} does not take into account the study of
the transient length. One can amend Lemma \ref{elll:nn1} to obtain the following
lemma:
\begin{lemma}  \cite{NT 03, NT 04} \label{llell:n2n} 
If there is a neuronal recurrence equation with memory length $k$
that generates a sequence $\{x^{\jmath}(n) : n \geq 0 \} \ ,  1
\leq \jmath \leq g$ of transient length $T_{\jmath}$ and of period
$p_{\jmath}$ then there is a neuronal recurrence equation with
memory length $kg$ that generates a sequence of transient length
$g \cdot max(T_1, T_2, \dots , T_g)$ and of period $g \cdot {\rm lcm}(p_1,
p_2, \dots, p_g).$
\end{lemma}
In the following example, we will show that Lemma \ref{elll:nn1} and Lemma \ref{llell:n2n} are incomplete. \\

{\bf Example 1:}  \\
Let us suppose that the neuronal recurrence equation defined by Equation (\ref{eeq:n1}) generated six sequences
\begin{equation}     \label{eqql:nn2}
\{ x^i(n) \ : \ n \ \geq 0    \}  \ , \ 0 \leq i \leq 5
\end{equation}
of periods
\begin{equation}     \label{eqql:nn4}
p_i \ = \ 1 \ , \ 0 \leq i \leq 5 \ 
\end{equation}
It is clear that each sequence defined by Equation (\ref{eqql:nn2}) is a fixed point. We present two different cases
 of evolution. \\
{\bf First case:}   \\
We suppose that 
\begin{align}
 x^{2i}(n)   &= 0 \ ; \forall n , i  \ {\rm such \ that} \ n \geq 0 \ {\rm and} \ 0 \leq i \leq 2    \label{eqql:nn8}  \\
 x^{2i+1}(n) &= 1 \ ; \forall n , i  \ {\rm such \ that} \ n \geq 0 \ {\rm and} \ 0 \leq i \leq 2    \label{eqql:nn9}         \end{align}
It is easy to verify that the shuffle of the neuronal recurrence equation defined by Equations (\ref{eqql:nn8}) and (\ref{eqql:nn9}) is  
\begin{multline}   \label{pmou:oupm90}
x^0(0)x^1(0) \dots x^5(0) x^0(1)x^1(1) \dots x^5(1) \cdots x^0(i)x^1(i) \dots x^5(i) \cdots   = \\
 010101010101010101010101 \cdots 010101010101 \cdots
\end{multline}
The sequence defined by Equation (\ref{pmou:oupm90}) describes a period of length $2$. By application of the Lemma \ref{elll:nn1} the period of the sequence defined by Equation (\ref{pmou:oupm90}) should be $6$ \ ( more precisely $6 \times lcm(1, 1, 1, 1, 1, 1)$ ). \\
{\bf Second case:}   \\
We suppose that 
\begin{align}
 x^{i}(n)   &= 0 \ , \forall n , i \ {\rm such \ that} \ n \geq 0 \ {\rm and} \ i \in \{ 0 , 1 , 3 , 4 \}   \label{eqql:nn10} \\
 x^{i}(n)   &= 1 \ , \forall n , i  \ {\rm such \ that} \ n \geq 0 \ {\rm and} \ i \in \{ 2 , 5 \}     \label{eqql:nn12}         \end{align}
It is easy to verify that the shuffle of the neuronal recurrence equation defined by Equations (\ref{eqql:nn10}) and (\ref{eqql:nn12}) is  
\begin{multline}  \label{pmou:oupm91}
x^0(0)x^1(0) \dots x^5(0) x^0(1)x^1(1) \dots x^5(1) \cdots x^0(i)x^1(i) \dots x^5(i) \cdots   =  \\ 
 001001001001001001001001001001001001 \cdots 001001 \cdots
\end{multline}
The sequence defined by Equation (\ref{pmou:oupm91}) describes a period of length $3$. By application of the Lemma \ref{elll:nn1} the period of the sequence defined by Equation (\ref{pmou:oupm91}) should be $6$ \ ( more precisely $6 \times lcm(1, 1, 1, 1, 1, 1)$ ).\\
The first case and the second case of example 1 show that Lemma \ref{elll:nn1} and Lemma 
\ref{llell:n2n} don't take into account all the cases. \\
One can amend the Lemma \ref{elll:nn1} as follows:
\begin{lemma} \label{elll:ss8} 
If there is a neuronal recurrence equation with memory length $k$ that generates $r$ sequences
of periods $p_1, p_2, \dots, p_r$, then there is a neuronal recurrence equation with memory length $kr$ that
generates a sequence of period $Per$. $Per$ is defined as follows: \\
{\it First case:} $\exists \ j, \ 1 \leq \ j  \ \leq \ r$ such that $p_j \geq 2$  \\
\[
Per \ = \ r  \times lcm(p_1, \cdots , p_r).
\]
{\it Second case:} $p_j = 1$ ;  $\forall \ j, \ 1 \leq \ j  \ \leq \ r$.  \\
\[
Per \ {\rm \ is \ a \ divisor \ of} \ r.
\]
\end{lemma}    

The improvement of Lemma \ref{elll:nn1} doesn't modify all the main results about periods obtained in the papers \cite{CTT 92, NM 00a, NM 00b, NT 04, TT 93} because all these main results consider only the case where the periods $p_1, p_2, \dots, p_r$ of the $r$ sequences are greater or equal to 2. \\
We can also amend the Lemma \ref{llell:n2n} as follows:
\begin{lemma}  \label{elnll:puon2}
If there is a neuronal recurrence equation with memory length $k$
that generates a sequence $\{x^{\jmath}(n) : n \geq 0 \} \ ,  1
\leq \jmath \leq g$ of transient length $T_{\jmath}$ and of period
$p_{\jmath}$ then there is a neuronal recurrence equation with
memory length $kg$ that generates a sequence of transient length
$g \cdot max(T_1, T_2, \dots , T_g)$ and of period $Per$. $Per$ is defined as follows: \\
{\it First case:} $\exists \ j, \ 1 \leq \ j  \ \leq \ r$ such that $p_j \geq 2$  \\
\[
Per \ = \ r  \times lcm(p_1, \cdots , p_g).
\]
{\it Second case:} $p_j = 1$ ;  $\forall \ j, \ 1 \leq \ j  \ \leq \ r$.  \\
\[
Per {\rm \ is \ a \ divisor \ of \ } g.
\]
\end{lemma}

\section{Preliminaries}

Let $k$ be a positive integer. For a vector $a \in \mathbb{R}^{k}$, a real
number $\theta \in \mathbb{R}$ and a vector $\phi \in \{0, 1 \}^{k}$. We
define the sequence $\{ x(n) \ : \  n \in \mathbb{N} \}$ by the following
recurrence:
\begin{equation}  \label{ch9:eq2}
x(t) = \begin{cases}
         \phi(t) \ ; \ & t \in \{  0, \dots , k-1 \}  \\
         {\bf 1} \Bigl( \sum_{i=1}^{k} a_{i}  x(t-i) - \theta \Bigl) \ ; \ & t \geq k \\
      \end{cases}
\end{equation}
We denote by $S(a, \theta, \phi)$ the sequence generated by equation
(\ref{ch9:eq2}), $Per(a, \theta, \phi)$ its period and $Tra(a, \theta, \phi)$ its transient length. \\

Let $m$ be a positive integer, we denote the cardinality of the
set ${\cal P} = \{ p \ : \ p \ prime \ {\rm and} \ 2m \ < \ p \ < \ 3m
\}$ by $\rho(m)$. Let us denote by $p_{0}, p_{1} , \dots , p_{-1+\rho(m)}$  the
prime numbers belonging  to  the  set  $\{ 2m+1, \ 2m+2, \dots, 3m-2, 3m-1 \}$,
 the sequence $\{ \alpha_i \ : \ 0 \leq i \leq -1+\rho(m) \}$ is defined as 
$\alpha_i \ = \ 3m-p_i , \ 0 \leq i \leq -1+\rho(m)$. \\
 We also suppose that:
\begin{equation} \label{pmou:oupm94}
p_{-1+\rho(m)} \ < \ p_{-2+\rho(m)} \ < \ \cdots \ < \ p_{i+1} \ < \ p_i \ < \ \cdots \ < \ p_1 \ < \ p_0  
\end{equation}
Subsequently, we consider only the integers $m$ such that $\rho(m) \geq 2$.  \\
It is easy to check that $\{ 2m+1, \ 2m+2, \dots, 3m-2, 3m-1 \}$ contains at most
$\lceil \frac{m-1}{2} \rceil$ odd integers. It follows that 
\begin{equation}  \label{ch:eq3}
  \rho(m) \ \leq \  \biggl\lceil \frac{m-1}{2} \biggl\rceil
\end{equation}
We set $k = (6m-1)\rho(m)$ and $\forall \ i \in \mathbb{N}, \ 0 \leq i \leq -1+\rho(m)$, 
we define :
\[
    \begin{aligned}
\mu(m, \alpha_i) &=  \biggl\lfloor \ \frac{k}{3m-\alpha_i} \  \biggl\rfloor  \\
   \beta(m, \alpha_i)   &=  k - ((3m-\alpha_i) \mu(m, \alpha_i))
    \end{aligned}
\]
From the previous definitions, we have $k=((3m-\alpha_i) \mu(m,\alpha_i))
+\beta(m,\alpha_i)$. \\
It is clear that $\forall \ i \ \in \ \mathbb{N}, \ 0 \leq i \leq -1+\rho(m)$
\[
2m+1 \ \leq \ 3m-\alpha_i \ \leq \ 3m-1
\]
This implies that
\[
\frac{(6m-1) \rho(m)}{3m-1}  \ \leq \ \frac{k}{3m-\alpha_i} \ \leq \ \frac{(6m-1) \rho(m)}{2m+1}
\]
Therefore
\begin{equation}  \label{zzr:rzz2}
2 \rho(m) \ \ \leq \ \ \mu(m, \alpha_i) \ \ \leq \ \ 3 \rho(m)
\end{equation}
$\forall \ i \in \mathbb{N}, \ 0 \leq i \leq -1+\rho(m)$, we want to construct a neuronal 
recurrence equation $\{ x^{\alpha_i}(n) \ : \ n \ \geq 0 \}$ with memory of 
length $k$ which evolves as follows :
\begin{equation}  \label{ch9:eq4}
\underbrace{00 \dots 0}_{\beta(m,\alpha_i)} \underbrace{100 \dots 0}_{3m-\alpha_i} 
\underbrace{100 \dots 0}_{3m-\alpha_i} \cdots \underbrace{100 \dots 0}_{3m-\alpha_i} \cdots  
\underbrace{100 \dots 0}_{3m-\alpha_i}  \cdots
\end{equation}
and which describes a cycle of length $3m-\alpha_i \ = \ p_i$. \\
$\forall \ i \in \mathbb{N}, \ 0 \leq i \leq -1+\rho(m)$, let $\phi^{\alpha_i} \ \in \ \{0,
1\}^{k}$ be the vector defined by
\begin{equation}  \label{ch9:eq5}
\phi^{\alpha_i}(0) \dots \phi^{\alpha_i}(k-1) = \underbrace{0 \dots 0}_{\beta(m, \alpha_i)} 
\underbrace{\underbrace{10 \dots 0}_{p_i} \cdots \underbrace{10 \dots 0}_{p_i}}_{\mu(m, \alpha_i)p_i }
\end{equation}
In other words, $\phi^{\alpha_i}$ is defined by:
\begin{equation}  \notag
     \phi^{\alpha_i}(j)  =
   \begin{cases}
1 &\text{if $\exists \ \ell , \ 0 \leq \ell \ \leq \mu(m, \alpha_i)-1$ such that $j \ = 
 \beta(m, \alpha_i)+ \ell p_i$} \notag \\
0 &\text{otherwise}  \notag
   \end{cases}
\end{equation}
We define the neuronal recurrence equation $\{ x^{\alpha_i}(n) \ : \ n \ \geq \ 0 \ \}$ by the following recurrence:
\begin{equation}  \label{ch9:eq6}
x^{\alpha_i}(t) = \begin{cases}
         \phi^{\alpha_i}(t) &\text{$ \ ; \ t \in \{  0, \dots , k-1 \}$}  \\
 {\bf 1} \Bigl( \sum_{j=1}^{k} \bar{a}_{j}  x^{\alpha_i}(t-j) - \bar{\theta} \Bigl)
            &\text{\ ; \ $ t \geq k$}
      \end{cases}
\end{equation}
where $\bar{a}_j$ is defined as follows:  \\

{\bf First case:} $\rho(m)$ is even and $\forall \ i_2 \in \mathbb{N}, \ 0 \leq i_2 \leq -1+\rho(m)$

\begin{equation}   \label{eeeqq:n2}
    \bar{a}_j  =
      \begin{cases}
2 &\text{if  $j \in Pos(\alpha_{i_2}) \ {\rm and} \  j \leq  \frac{3 \times \rho(m) \times p_{i_2}}{2} \ ,$}  \\
-2 &\text{if $j \in Pos(\alpha_{i_2}) \ {\rm and} \ j \ >  \frac{3 \times \rho(m) \times p_{i_2}}{2} \ , $}  \\
 0  &\text{otherwise. }
      \end{cases}
\end{equation}

{\bf Second case:} $\rho(m)$ is odd, $\rho(m) \geq 3$ and $\forall \ i_2 \in \mathbb{N}, \ 0 \leq i_2 \leq -1+\rho(m)$ 

\begin{equation}   \label{eeq:n333}
     \bar{a}_j  =
      \begin{cases}
 2 &\text{if $j \in Pos(\alpha_{i_2}) \ {\rm and} \ j \leq  \frac{(3 \rho(m)-1)}{2} \times p_{i_2} \ , $ } \\
-2  &\text{if $j \in Pos(\alpha_{i_2}) \ {\rm and} \ \frac{(3 \rho(m)+1)}{2} \times p_{i_2}  \leq j \leq (2 \rho(m)-2) \times p_{i_2}, $} \\
-1 &\text{if $j \in \{(2\rho(m)-1) \times p_{i_2} \ , \ 2 \rho(m) \times p_{i_2} \}$ \ , }   \\
 0  &\text{otherwise. }
      \end{cases}
\end{equation}

We also define:

\begin{align}
 Pos(\alpha_i) &= \{ jp_i : j = 1, \dots , 2\rho(m) \}  \label{pp1:nn1}   \\
       &= \{ p_i, 2p_i, \dots , (-1+ 2\rho(m))p_i , 2\rho(m)p_i \}, 
     \ \ 0 \ \leq i \ \leq -1+\rho(m)   \label{pp1:nn2}   \\
 D &= \{ i : i = 1, \dots, k \} = \{ 1, 2, \dots , k-1, k \}  \label{pp1:nn3}    \\
 F &=   \bigcup_{i=0}^{-1+\rho(m)} Pos( \alpha_i)    \label{pp1:nn4}     \\
 G &=  D \setminus F     \label{pp1:nn5}     \\
\bar{\theta} &= \ 2 \times \rho(m) \label{pp1:nn6}  \\
 k &= (6m-1) \times \rho(m)     \label{pp1:nn7} 
\end{align}
By definition $Pos(\alpha_i)$ represents the set of indices $j, \ 1 \leq j \leq k$ such that 
$x^{\alpha_i}(k-j) \ = \ 1.$ \\
From the definition of $Pos(\alpha_i)$ and from Equation (\ref{ch9:eq5}), one
can easily verify that
\begin{align}  
j \ \in \ Pos(\alpha_i) \ &\Longrightarrow \ x^{\alpha_i}(k-j) \ = \ 1 \ 
                        \label{ch9:eq7}     \\
j \ \in \ D \setminus Pos(\alpha_i) \ &\Longrightarrow \ \ x^{\alpha_i}(k-j) \ = \ 0 \ \
                        \label{ch9:eq8}    
\end{align}
$\forall \ d \ \in \ \mathbb{N}, \ 0 \ < \  d \ < \ p_i$, we also denote $PPos(\alpha_i, d)$ the 
set of indices $j$ such that $x^{\alpha_i}(k+d-j) \ = \ 1$, in other words:
\[
PPos(\alpha_i, d) \ = \ \{ \ j : x^{\alpha_i}(k+d-j) = 1 \ {\rm and} \ 1 \leq j \leq k \ \}
\]
$\forall \ i, d \ \in \ \mathbb{N}, \ 0 \ \leq  i \ \leq -1+\rho(m)$ and  $0 \ < \ d \ < \ p_i$, we denote:
\begin{align}
Q(\alpha_i, d)  &= \{ d+jp_i : j = 0, 1, \dots , \mu(m, \alpha_i) \}, \ \ \ 0 < d \leq \beta(m, \alpha_i)  \notag  \\
Q(\alpha_i, d)  &= \{ d+jp_i : j = 0, 1, \dots , -1+\mu(m, \alpha_i) \}, \ \ \beta(m, \alpha_i) < d < p_i \notag  \\
E(\alpha_i, d)  &=  Q(\alpha_i, d) \ \cap \ F   \notag 
\end{align}
The neuronal recurrence equation $\{ x^{\alpha_i}(n) \ : \ n \geq 0 \}$ with memory of 
 length $k$ is defined by Equations (\ref{ch9:eq5}) and (\ref{ch9:eq6}). \\
We will show that the neuronal recurrence equation $\{ x^{\alpha_i}(n) \ : \ n \geq 0 \}$ 
evolves as specified in equation (\ref{ch9:eq4}). \\
In the following proposition, we present an important property.
\begin{proposition}  \label{ch9:prop1}  \cite{NT 04}
$\forall \ i \ \in \ \mathbb{N}, 0 \leq i \leq -1+\rho(m)$  and
$\forall \ d \ \in \ \mathbb{N}, 1 \leq d < p_i$   \\
\begin{eqnarray}    \notag
  card \ E(\alpha_i, d)  \leq  \rho(m) - 1.
\end{eqnarray}
\end{proposition}

The following proposition characterizes the sum of the interaction coefficients $\bar{a}_j$ when $j \in Pos(\alpha_i)$ 
\begin{proposition}  \label{ch9:prop22}  
$\forall \ i \ \in \ \mathbb{N}, 0 \leq i \leq -1+\rho(m)$, we have
\begin{eqnarray}    \notag
\sum_{j \in Pos(\alpha_i)}  \bar{a}_j \ = \ 2 \times \rho(m).
\end{eqnarray}
\end{proposition}

The following lemma characterizes the evolution of the sequence
$\{ x^{\alpha_i}(n) \ : \ n \geq 0 \}$ at time $t \ = \ k$.
\begin{lemma}   \label{ch9:lem1}
\[
x^{\alpha_i}(k) = 1.
\]
\end{lemma}

From Lemma \ref{ch9:lem1} and Equation (\ref{ch9:eq5}), it is easy to verify that
\begin{equation}   \label{ch9:eq30}
PPos(\alpha_i, 1) \ = \ Q(\alpha_i, 1)
\end{equation}
From the definition of $E(\alpha_i, 1)$, from Equation (\ref{ch9:eq5}), from 
Equation (\ref{ch9:eq30}) and from the Lemma \ref{ch9:lem1}, we check easily that:
\begin{equation}  
 \ell \ \in \ D \setminus E(\alpha_i, 1) \ \ \Longrightarrow \ \ x^{\alpha_i}(k+1-\ell) = 0 \ 
      \ {\rm or} \ \ \bar{a}_{\ell} \ = \ 0.    \ \ \label{ch12:eqq2}
\end{equation}
The values of the sequence $\{ x^{\alpha_i}(n) \ : \ n \geq 0 \}$ at time 
$t = k+1 , \dots , k-1+p_i$ are given by the following lemma.
\begin{lemma}   \label{ch9:lem2}
\[
\forall \ t \ \in \ \mathbb{N} \ {\rm such \ that} \ 1 \leq t
\leq 3m-1- \alpha_i, \ {\rm we \ have \ } \ x^{\alpha_i}(k+t) =
0.
\]
\end{lemma}

It is easy to verify that $\forall \ i \ \in \mathbb{N}, \ 0 \ \leq \ i  \leq \ -1+\rho(m)$, we have:
\[
PPos(\alpha_i, j) \ = \ Q(\alpha_i, j)  \  \  \forall \ j, \  \  1 \  \leq  \  j  \  \leq  3m-1-\alpha_i
\]
\begin{lemma}  \label{ch9:lem3}
\[
{\rm There \ exists} \ \bar{a}, \phi^{\alpha_i} \in \mathbb{R}^{k} \ and \ \bar{\theta} \in \mathbb{R} \ {\rm such \ that \ :}
\]
\[
Per(\bar{a}, \bar{\theta}, \phi^{\alpha_i} ) = p_i.
\]
\end{lemma} 
\bigskip
\begin{lemma}  \label{ellll:n3} 
$\forall \ t, i \in \mathbb{N}$,  $t \geq k$ and $0 \leq i \leq -1+\rho(m)$ \\
\[
\mu(m,\alpha_i) \ \leq \sum_{j=1}^{k}  x^{\alpha_i}(t-j) \ \leq 1+\mu(m,\alpha_i).
\]
\end{lemma}

In order to present some properties of the sequence $\{x^{\alpha_i}(n) : n \geq 0 \}$, 
we introduce the following notation:
\begin{notation}
Let us define $S1(\alpha_i, n)$ as:
\[
S1(\alpha_i, n) =  \sum_{j = 1}^{k} \bar{a}_j x^{\alpha_i}(n-j)  
\]
and let $\lambda$ be a strictly negative real number such that: $\forall \ i, \ 0 \leq i \leq \rho(m)-1$
\[
\ \ max \ \{ \ S1(\alpha_i, n) - \bar{\theta} \ : \ S1(\alpha_i ,n) \ < \ \bar{\theta} \ and \ n \geq k \} \leq \lambda
\]
\end{notation}
\bigskip

\begin{lemma}  \label{ell:n4}
$\forall \ i, n \in \mathbb{N}$ such that $0 \leq i \leq -1+\rho(m)$ and
$ n \geq k$,
\[
S1(\alpha_i, n) \ \in \ [ -2(1+\mu(m, \alpha_i)) , \bar{\theta} - 1 ] \cup \{ \bar{\theta} \} ,
\]
\[
\lambda \ \in [-1, 0 [.
\]
\end{lemma}

Let $\{ v^{\alpha_i}(n) : n \geq 0 \}$ be the sequence whose first $k$ terms
are defined as follows:
\begin{equation}  \label{eeq:n7}
v^{\alpha_i}(0) v^{\alpha_i}(1) \dots v^{\alpha_i}(k-1) \ = \ x^{\alpha_i}(1) \cdots x^{\alpha_i}(k-1) \overline{x^{\alpha_i}(k)},
\end{equation}
and the other terms are generated by the following neuronal
recurrence equation:
\begin{equation}   \label{eeq:n8}
v^{\alpha_i}(n) = {\bf 1} \left[ \sum_{j=1}^{k} \bar{a}_{j} v^{\alpha_i}(n-j)- \bar{\theta} \right] , \ \ \ n \geq k.
\end{equation}

\begin{remark} 
The term ${x^{\alpha_i}(k)}$ is equal to $1$, this implies that $v^{\alpha_i}(k-1)$ is equal to $0$. 
\end{remark}

The parameters $\bar{a}_j, 1 \leq j \leq k$ and $\bar{\theta}$ used in
neuronal recurrence Equation (\ref{eeq:n8}) are those defined in Equations (\ref{eeeqq:n2}), (\ref{eeq:n333}) and (\ref{pp1:nn6}).\\

The following lemma, which is easy to prove, characterizes the
evolution of the sequence $\{ v^{\alpha_i}(n) : n \geq 0 \}$.
\begin{lemma} \label{ell:n5}
In the evolution of the sequence $\{v^{\alpha_i}(n) : n \geq 0 \}$, $\forall \ t \ \in \mathbb{N}, \ t \geq k$ we have: \\
(a) $v^{\alpha_i}(t) = 0$,  \\
(b) $\sum_{j=1}^{k} \bar{a}_{j} v^{\alpha_i}(t-j) \ \leq \ \bar{\theta}-2$, \\
(c) The sequence $\{ v^{\alpha_i}(n) : n \geq 0 \}$ describes a transient of length $k-p_i$ and a fixed point. 
\end{lemma}

The instability of the sequence $\{ x^{\alpha_i}(n) : n \ \geq \ 0 \}$
occurs as a result of the convergence of the sequence $\{ v^{\alpha_i}(n) : n
\ \geq \ 0 \}$ to $0 \ 0 \cdots 0 \ 0$. 
\begin{notation}.  
$h \ = \ \rho(m) \times k = (6m-1) \times (\rho(m))^2$ is the length of the memory of some neuronal recurrence equations.
\end{notation}

Let us also note:
\begin{equation}   \label{qeqq:n50}
      L_0(d) =
      \begin{cases}
      \rho(m) \times lcm(p_{d+1}, p_{d+2}, \dots , p_{-2+\rho(m)}, p_{-1+\rho(m)}) \ ,&\text{if $0 \leq d \leq -2+\rho(m)$ } \\
      1 \ ,   &\text{if $d = -1+\rho(m)$.} 
      \end{cases}
\end{equation}
\begin{align} 
L_1(d) &= \rho(m) \times lcm(p_{0}, p_{1}, \dots , p_{d}) \ , \ 0 \leq d \leq \rho(m)-1 \label{qeqqn:nn52}   \\
L_2    &=  \rho(m) \times lcm(p_{0}, p_{1}, \dots , p_{-1+\rho(m)}) \label{qeqqn:nn53}
\end{align}
$L_0(d)$, $L_1(d)$ and $L_2$ represent the periods of some neuronal recurrence equations.   \\

Let $\{ y(n)  : n \geq 0 \}$ be the sequence whose first $h$ terms are defined as follows: 
\begin{equation} \label{eeq:n9}
\forall \ j \in \mathbb{N}, \ 0 \leq j \leq \ k-1 \ \ \
y\left((\rho(m) \times j)+i \right) = x^{\alpha_i}(1+j), \ \  0 \leq  i \leq -1+\rho(m)
\end{equation}
and the other terms are generated by the following neuronal recurrence
equation:
\begin{equation}  \label{eeq:n10}
y(n) = {\bf 1} \left[ \sum_{f=1}^{h} b_{f}y(n-f)- \theta_1 \right]  \ ; \ n \geq h
\end{equation}
where
\begin{equation}   \label{eeq:n11}
      b_{f} =
      \begin{cases}
      \bar{a}_{j} \ ,   &\text{if $f = \rho(m) \times j, \ \ 1 \leq j \leq k$ } \\
      0 \ ,   &\text{otherwise.} 
      \end{cases}
\end{equation}
\begin{equation} \label{eeq:n12}
\theta_1 = \bar{\theta}.
\end{equation}
The parameters $\bar{a}_j$ are those defined in Equations (\ref{eeeqq:n2}) and (\ref{eeq:n333}). The parameters
$\bar{\theta}$ and $k$ are defined in Equations (\ref{pp1:nn6}) and (\ref{pp1:nn7}).
\begin{remark} \label{r:n1}  
(a) The first $h$ terms of the sequence $\{ y(n) : n \geq 0 \}$
are obtained by shuffling the $k$ terms of each subsequence
$\{x^{\alpha_i}(n) : 1 \leq n \leq k \}$ where $0 \leq i \leq -1+\rho(m)$. \\
(b) The neuronal recurrence equation (\ref{eeq:n10}) is obtained by
applying the construction of Lemma \ref{elll:nn1} to the neuronal recurrence
Equation (\ref{ch9:eq6}) whose parameters are given in
Equations (\ref{eeeqq:n2}), (\ref{eeq:n333}), (\ref{pp1:nn6}) and (\ref{pp1:nn7}). \\
\end{remark}
From the fact that the sequence $\{ y(n) :
n \geq 0 \}$ is the shuffle of the $\rho(m)$ subsequences and from
its construction, we can write:
\begin{lemma}  \label{ell:n6}
$\forall \ t \ \in \mathbb{N}$ such that $t \ = q \times \rho(m) +i$ with
$q \ \in \mathbb{N}$ and $0 \leq i \leq -1+\rho(m)$, we have:
\[
y(t) \ = \ x^{\alpha_i}(1+q).
\]
\end{lemma}
The next lemma gives the period of the sequences $\{ y(n) : n \geq 0 \}$.
\begin{lemma}  \label{ell:n7}
The sequence $\{ y(n) \ : \ n \geq 0 \}$ describes a cycle of length $L_2$.
\end{lemma}

$\forall \ d \in \mathbb{N}$ such that $0 \leq d \leq -1+\rho(m)$, we denote by $\{ w(n, d) : n \geq 0 \}$ the sequence whose first $h$ terms are defined as:
\begin{eqnarray}  \label{ComNov:n09}
\forall \ i, \ 0 \leq i \leq d \ \ \ w(\rho(m)j+i, d) =
      \begin{cases}
x^{\alpha_i}(1+j), \  &\text{$0 \leq \ j \ \leq \ k-2$} \\
\overline{x^{\alpha_i}(k) } \ ,         &\text{j = k-1.}
      \end{cases}
\end{eqnarray}
and
\begin{eqnarray}  \label{ComNov:n19}
\forall \ i, \ d+1 \leq i \leq -1+\rho(m) \ \ w(\rho(m)j+i, d) = y(\rho(m)j+i+L_1(d)) \ ; \ 0 \leq \ j \ \leq \ k-1
\end{eqnarray}
The first $h$ terms of the sequence $\{ w(n, d) : n \geq 0 \}$ are
obtained by shuffling the $k$ terms of each of the sequences:
\begin{equation}   \label{eeq:n13}
v^{\alpha_i}(0) \ v^{\alpha_i}(1) \ v^{\alpha_i}(2) \dots  v^{\alpha_i}(k-1) \ ; \ 0 \leq i \leq d
\end{equation}
and
\begin{equation}  \label{eeqqe:n2n25}
x^{\alpha_i}(1+\gamma_i(d)) \ x^{\alpha_i}(2+\gamma_i(d)) \ x^{\alpha_i}(3+\gamma_i(d)) \dots  x^{\alpha_i}(k+\gamma_i(d)) \ ;  \ d+1 \leq i \leq -1+\rho(m)
\end{equation}
where:
\begin{eqnarray} \label{eqn11c0}
\frac{L_1(d)}{\rho(m)}  \equiv \gamma_i(d) \pmod{p_i} \ ; \ d+1 \leq i \leq -1+\rho(m).
\end{eqnarray}
The other terms of the sequence $\{ w(n, d) : n \geq 0 \}$ are
generated by the following neuronal recurrence equation:
\begin{equation}   \label{eeq:n15}
w(n, d) = {\bf 1} \left[ \sum_{f=1}^{h} b_{f} w(n-f, d)- \theta_1 \right] \ ; \ n \geq h
\end{equation}
The next lemma gives the period of the sequence $\{ w(n, d) : n \geq 0 \}$.
\begin{lemma}  \label{ell:n9}
The sequence $\{ w(n, d) : n \geq 0 \}$ generates a transient of length $\left(\rho(m) \times (k-p_{d}-1) \right)+d+1$ and a cycle of length $L_0(d)$.
\end{lemma}

\begin{notation}
Let us define $S2(n), \ S3(n, d)$ as:
\begin{eqnarray}   \notag
S2(n)           &=&  \sum_{f = 1}^{h} b_{f} y(n-f),        \notag    \\
S3(n, d)        &=&  \sum_{f = 1}^{h} b_{f} w(n-f, d).      \notag   
\end{eqnarray}
\end{notation}
\begin{remark}  \label{r:n2}
On the basis of the composition of automata \cite{CTT 92} and the
definition of $\lambda$ , we can conclude that:
\begin{list}{\texttt{$\bullet$}}{}
\item  ${\rm max} \ \{ \ S2(n) - \theta_1 \ : \ S2(n) \ < \ \theta_1 \ {\rm
and} \ n \geq h \} = {\rm max} \ \{ \ S1(r, n) - \bar{\theta} \ : \
S1(r, n) \ < \ \bar{\theta} \ and \ n \geq k \}  \ \leq \lambda$, and  
\item  ${\rm max} \ \{ \ S3(n, d) - \theta_1 \ : \ S3(n, d) \ < \ \theta_1 \ {\rm
and} \ n \geq h \} = {\rm max} \ \{ \ S1(r, n) - \bar{\theta} \ : \
S1(r, n) \ < \ \bar{\theta} \ and \ n \geq k \}  \ \leq \lambda$.
\end{list}
\end{remark}
\subsection{Results on the dynamics of sequences $y$ and $w$}
In this subparagraph, we recall and give some interesting results
on dynamics of the sequences $\{ y(n) : n \geq 0 \}$ and 
$\{ w(n, d) : n \geq 0 \}$. \\
The following lemma characterizes the sequence $\{ y(n) : n \geq 0
\}$ and $\{ w(n, d) : n \geq 0 \}$ in terms of the sum of $h$ consecutive
terms.
\begin{lemma} \label{ell:n8}
$\forall \ n, d \in \mathbb{N}$,such that $n \geq h$ and $0 \leq d \leq \rho(m)-1$ 
\begin{list}{\texttt{$\bullet$}}{}
\item $\sum_{i=0}^{-1+\rho(m)} \mu(m, \alpha_i) \ \leq \sum_{f=1}^{h}  y(n-f) \ \leq \rho(m) + \sum_{i=0}^{-1+\rho(m)} \mu(m, \alpha_i)$,
\item $\sum_{i=d+1}^{-1+\rho(m)} \mu(m, \alpha_i) \ \leq \sum_{f=1}^{h} w(n-f, d) \ \leq -d-1+\rho(m)+\sum_{i=0}^{-1+\rho(m)} \mu(m, \alpha_i)$.
\end{list}
\end{lemma}
 
\begin{definition}
Let us define the terms of the sequence $\{ tim(i,l, q) \ : l, q \in \mathbb{N} \
{\rm and} \ 0 \leq i \leq -1+\rho(m) \}$ as: \\
\[
tim(i,l,q) = (l \times \rho(m) ) + i + q \  :  \ \ l, q \ \in \mathbb{N} \ {\rm and} \ 0 \leq i \ \leq -1+\rho(m).
\]
\end{definition}
From Lemma \ref{ell:n6}, one can easily deduce that:
\begin{multline}  \label{eeq:n16}  
y(tim(i,l,0)) y(tim(i,l+1,0)) \dots y(tim(i,l+k-1,0)) =  \\  x^{\alpha_i}(1+l) \dots \
x^{\alpha_i}(l+k) \ ; \ 0 \leq i \leq -1+\rho(m).
\end{multline}
From Lemma \ref{ell:n6} or Equation (\ref{eeq:n16}), we also easily deduce 
that the terms of the sequence $\{ y(n) : n \geq 0 \}$ verify the following 
relation:

\begin{subequations}  \label{ren:nd9}
\begin{multline}  
y(tim(i,0,-\rho(m)+L_1(d))) y(tim(i,1,-\rho(m)+L_1(d))) \dots y(tim(i,k-1,-\rho(m)+L_1(d))) = \\
\underbrace{00 \dots 0}_{\beta(m, \alpha_i)} \underbrace{\underbrace{10
\dots \ 0}_{p_i} \dots \underbrace{10 \dots \ 00}_{p_i}}_{\mu(m, \alpha_i) \times p_i} \ ; \ 0
\leq i \ \leq d \
\end{multline}

\begin{multline} 
y(tim(i,0,-\rho(m)+L_1(d))) y(tim(i,1,-\rho(m)+L_1(d))) \dots y(tim(i,k-1,-\rho(m)+L_1(d))) = \\
x^{\alpha_i}(\gamma_i(d)) x^{\alpha_i}(1+\gamma_i(d)) \dots \  x^{\alpha_i}(k-2+\gamma_i(d)) x^{\alpha_i}(k-1+\gamma_i(d)) ; \ d+1 \leq i \ \leq -1+\rho(m). 
\end{multline}
\end{subequations}

For all $d \in \mathbb{N}$ such that $0 \leq d \leq -1+\rho(m)$, we note $B_0(d)$ the set of integers $f$ which verify Equations (\ref{ren:nd2}) and (\ref{ren:nd4}):
\begin{equation} \label{ren:nd2} 
1 \leq f \leq h-d 
\end{equation}
\begin{equation} \label{ren:nd4}
y(-f-\rho(m)+h+L_1(d)) \ = \ 1.
\end{equation}

{\bf Comment :}   By considering the following terms:  $y(-\rho(m)+L_1(d)) \ y(1-\rho(m)+L_1(d)) \dots y(-\rho(m)+L_1(d)+h-1)$, it is 
 possible by easy computation to build the set $B_0(d)$. In a bid  to give the algebraic expression of the set $B_0(d)$, let us define the set $C0(i, p_i)$ 
 as follows:
 
\begin{equation}   \label{v4ren:t0}
C0(i, p_i)  =    \left\{\,  -i + \rho(m) + ( \rho(m)  \times  p_i \times  j )  \  :  \  j = 0, 1 , \dots , \mu(m,  \alpha_i)   \    \,\right\}   \    ,   0   \leq  i  \leq  \rho(m)-1  .
\end{equation}
From the definition of the terms $y(0)y(1)  \dots  y(h-1)$ (see Equation (\ref{eeq:n9})) , it is easy to see that:  $\bigcup_{i=0}^{\rho(m)-1} C0(i, p_i)$ represents the set of indices $j$ such that
$y(h-j) = 1$.    \\
Let us define the set $C1(n, y)$ as follows:
\begin{equation}   \label{v4ren:t1}
C1(n , y )  =    \left\{\,   j    \   :  \    y(h+n-j) =  1   \,\right\}  
\end{equation}
It is easy to see  that:
\begin{align}   \notag
 \ell   \in \bigcup_{i=0}^{\rho(m)-1} C0(i, p_i)     &\Longleftrightarrow   y(h- \ell) = 1   \ {\rm   and}   \   1   \   \leq   \ell  \   \leq  h,          \notag     \\
 		                                                                &\Longleftrightarrow   y(h+n - (\ell+n) ) = 1   \ {\rm   and}   \   1   \   \leq   \ell  \   \leq  h,         \notag    \\
                                                                                   &\Longleftrightarrow   \ell + n   \in   C1( n ,  y )   \   {\rm   and}   \   1   \   \leq   \ell  \   \leq  h.   \notag 
\end{align}
Consequently:
\begin{equation}   \label{v5ren:t3}
\ell  \   \in  \  \bigcup_{i=0}^{\rho(m)-1} C0(i, p_i)    \Longleftrightarrow   \ell + n  \  \in  \  C1(n,  y)  \ {\rm   and}   \   1   \   \leq   \ell  \   \leq  h,       
\end{equation}
Based on the period of the subsequence $\{x^{\alpha_i}(n)  \  :   \  n \geq  0 \}$ and on the fact that $y$ is  a shuffle of the subsequences $x^{\alpha_i}   \  (  \  0   \leq  i  \leq   \rho(m)-1  \  )$,
 we deduce that:
 
\begin{displaymath}   
\left.
       \begin{array}{ll}
    \ell+n   \  \in  \  C1(n ,  y)   ,       & \hbox{} \\
    \ell        \  \in  \  C0(i ,  p_i)     ,  & \hbox{} 
        \end{array}
\right \}                        \Longrightarrow      \ell +n  - ( \rho(m)  \times   p_i )  \in  \  C1(n ,  y).
\end{displaymath}
 
Let us  denote $C2(i, d)$ the following set: 
\begin{equation}   \label{v6ren:t5}
C2(i, d)  =    \left\{\,    j   \   :   \   -i +  \rho(m) + L_1(d)-    \rho(m)    \equiv  j  \pmod{ \rho(m)  \times p_i  }    \,\right\}.  
\end{equation}
The set $C2(i, d)$ contains the only element  $j$  ( $0  \leq  j  \  <   \rho(m)  \times  p_i$  ) such that  $-i +  \rho(m) + L_1(d) -  \rho(m)   \equiv  j  \pmod{ \rho(m)  \times p_i }$. By  using the Equation (\ref{v5ren:t3})  and the fact that   $-i +  \rho(m)  \   \in  \  C0(i ,  p_i)$, we deduce  the following  implications:
 
\begin{displaymath}                                                                                                   
\left.
\begin{array}{lll}
   -i +  \rho(m)  \   \in  \  C0(i ,  p_i),  & \hbox{} \\
     j   \  \in  \  C2(i ,  d),                       & \hbox{} \\
     j   \ne  0,                                         & \hbox{}
              \end{array}
\right \}
\Longrightarrow     j   \   \in  \  C1(n ,  y)                   
\end{displaymath}
   
\begin{displaymath}   
\left.
\begin{array}{lll}
   -i +  \rho(m)  \   \in  \  C0(i ,  p_i),  & \hbox{} \\
     j   \  \in  \  C2(i ,  d),                       & \hbox{} \\
     j    =   0,                                          & \hbox{}
              \end{array}
\right \}
\Longrightarrow    \rho(m)   \times  p_i    \    \in  \  C1(n ,  y)                   
\end{displaymath}

We  build the set  $C3(i, d)$ as  follows:
\begin{equation}   \label{v8ren:t6}
C3(i, d)  =    \left\{\,   \ell    \   :   \    \ell     \equiv  j  \pmod{ \rho(m)  \times p_i  }     \    ,  \   1   \leq   \ell    \leq  h   \   ,   \    \ell   \  \in  \  C2(i, d)    \,\right\}.  
\end{equation} 
It is easy to see  that:
\begin{equation}   \notag
B_0(d)  \   =   \  \bigcup_{i=0}^{\rho(m)-1} C3(i,  d).       
\end{equation}

Let us denote $A(d)$ the following set:
\begin{align}  
 A(d)             &= \bigcup_{i=0}^{d} B_i(d)  \ {\rm where}     \label{nre:n2d2}  \\
B_{i+1}(d)   &= \left\{\ 1+\jmath \ : \ \jmath \ \in B_i(d) \ \right\}    \notag 
\end{align}
Based on:
\begin{itemize}
 \item the evolution of the neuronal recurrence equation $y$,
 \item on the definition of the set $B_{i}(d)$,
\end{itemize}
it is easy to verify that:
\begin{equation} \label{ren:nd14}
\forall \ell \in \mathbb{N}, \ 0 \leq \ell \leq d \ {\rm and \ } \forall f \in B_{\ell}(d) \ {\rm we \ have \ } y(-f-\rho(m)+h+\ell+L_1(d)) \ = \ 1.
\end{equation}

\section{Construction of the neuronal recurrence equation $z$ }
The basic idea is to construct a sequence $\{ z(n, d) : n \geq 0 \}$
whose terms are generated by the following neuronal recurrence
equation:
\begin{equation}   \label{eeq:n21} 
z(n, d) = {\bf 1} \left[ \sum_{f=1}^{h} c(f, d) z(n-f, d)- \theta_2(d) \right] ,
\end{equation}
and whose  first $h$ terms are initialized as follows:
\begin{equation}  \label{eeq:n22}
z(f, d) =  y(f) \ \ ; \ \  0 \leq f \leq h-1
\end{equation}
which exploits the instability of the sequences $\{x^{\alpha_i}(n) : n \
\geq 0 \}$ to converge to the sequence $\{ w(n,d) \ : \ n \geq 0 \}$. \\
We define the parameters $c(f,d)$ and $\theta_2(d)$ as follows:
\begin{equation}  \label{eeq:n23}
 c(f, d) =
\begin{cases}
b_{f}          &\text{ $1 \leq f \leq h \ and \ f \notin A(d)$ , } \\
b_{f} + \beta(d)  &\text{ $1 \leq f \leq h \ and \ f \in A(d)$. }
\end{cases}
\end{equation}
\begin{equation}  \label{eeq:n24}
\theta_2(d) = \theta_1 + \xi(d),
\end{equation}
where:
\begin{align} \label{eeq:n25}
\beta(d)  &= \frac{\lambda}{Tot(d)} ,  \\ 
\xi(d)    &=  \lambda - \frac{\beta(d)}{8} ,   \\ 
Tot(d)    &= card \ B_{\ell}(d) , \ 0 \leq \ell \leq d.
\end{align}
We have defined the parameters $c(f, d)$ so
that the sequence $\{ z(n, d) : n \geq 0 \}$ converges to the sequence $\{ w(n, d) : n \geq 0 \}$. \\
\begin{remark} \label{r:n3}
The terms of the sequence $\{ z(n, d) : n \geq 0 \}$
verify the following relation:
\begin{multline}   \notag
z(tim(i,0,0), d) z(tim(i,1,0), d) \dots z(tim(i,k-1,0), d) =  \\ x^{\alpha_i}(1)
x^{\alpha_i}(2) \dots \  x^{\alpha_i}(k-1) x^{\alpha_i}(k) \ \ ;
\ \ 0 \leq i \leq  -1+\rho(m).
\end{multline}
\end{remark}
\bigskip
\begin{notation}
Let $Q2(n, d)$ be defined as follows:
\begin{eqnarray}
Q2(n, d) = \beta(d) \sum_{f \in A(d)} z(n-f, d) - \xi(d).  \notag
\end{eqnarray}
\end{notation}
We establish below a proposition which states a relation between
$z(n, d)$, $S2(n)$ and $Q2(n, d)$.
\begin{proposition} \label{epp:n1} 
{\rm If} 
\[
z(n-i,d) \ = \ y(n-i) \ {\rm for \ all \ } i {\rm \ such \ that \ } 1 \leq i \leq h \ {\rm and} \ n-i \equiv 0 \pmod{\rho(m)}
\]
{\rm Then} 
\[ 
z(n, d) = {\bf 1} \left[ S2(n) + Q2(n, d)- \theta_1 \right].
\]
\end{proposition}

Below, we establish two propositions which provide the link
between the value  of $Q2(n, d)$ and the sequence $z(n, d)$.
\begin{proposition}  \label{epp:n2}
{\rm If} 
\[
\sum_{f \in A(d)} z(n-f, d) =  Tot(d)
\]
{\rm Then}
\[
Q2(n, d) = \frac{\beta(d)}{8}.
\]
\end{proposition}

\begin{proposition}  \label{epp:n3}
{\rm If} 
\[
\sum_{f \in A(d)} z(n-f, d) \leq  Tot(d)-1
\]
{\rm Then}
\[
\frac{-7 \beta(d)}{8} \leq  Q2(n, d)  \leq - \xi(d).
\]
\end{proposition}

In the next two lemmas, we show the relations between $B_0(d)$, $A(d)$ and $Tot(d)$. 
\begin{lemma} \label{lem:mel88}
$\forall d, n \in \mathbb{N}$ such that $0 \leq d \leq \rho(m)-1$   \\
{\bf If}  \\
\[
 n \not\equiv h-\rho(m)+L_1(d) \pmod{L_2}
\]
{\bf Then}
\[
 \sum_{f \in B_0(d)} y(n-f)  \ < \ Tot(d).
\]
\end{lemma}

\begin{lemma}  \label{lem:mel99}
$\forall n \in \mathbb{N}$ such that $n \equiv \ell \pmod{L_2}$  \\
{\bf If}  \\
\[
\ell \not\in  \{ -\rho(m)+h+L_1(d), 1-\rho(m)+h+L_1(d), \dots, d-\rho(m)+h+L_1(d) \}
\]
{\bf Then}
\[
 \sum_{f \in A(d)} y(n-f)  \ < \ Tot(d).
\]
\end{lemma}

We want to exploit the fact that:
\begin{list}{\texttt{$\bullet$}}{}
\item  $\sum_{f \in A(d)} z(n-f, d) = Tot(d)$ implies that $Q2(n, d) = \frac{\beta(d)}{8} \ < \ 0.$  \\
\item  $\sum_{f \in A(d)} z(n-f, d) \leq Tot(d)-1$ implies that $\frac{-7 \beta(d)}{8} \leq Q2(n, d) \leq -\xi(d).$ \\
\item $\forall t, \ 0 \leq t \ \leq \ \rho(m)-1$ we have $\sum_{f=1}^{h} b_f y\left(h-\rho(m)+t-f+L_1(-1+\rho(m))\right) = \bar{\theta}.$
\item the instability of the sequence $\{ x^{\alpha_i}(n) : n \geq 0 \}.$
\end{list}
to prove that the sequence $\{ z(n, d) : n \geq 0 \}$ converges to the sequence $\{ w(n, d) : n \geq 0 \}$.
We intend to divide the dynamic of the sequence $\{ z(n, d) : n \geq 0 \}$ into five phases.

\section{Dynamical behavior of the neuronal recurrence equation $z$}
This study is done in two steps : first we analyse the
transient phase and next the cyclic phase. Subsequently, we suppose that $d$ verifies the following equation
\[
d \in \mathbb{N} \ {\rm and \ } \ 0 \leq d \leq \rho(m)-1
\]
\subsection{Transient Phase }
The transient phase of the sequence $\{ z(n,d): n \geq 0 \}$ unfolds
during four phases. \\
{\bf Phase 1.} We want that the dynamics of the sequence $\{ z(n, d)
: n \geq 0 \}$ verifies the relation: $\forall \ t \in \mathbb{N}$
such that $0 \leq t \leq L_1(d)+h-1-\rho(m)$, we have:
\[
z(t, d) =  y(t)
\]
Phase 1 starts at time 0 and finishes at time $L_1(d)+h-1-\rho(m)$. \\
In the lemma below, we prove that the sequence $\{ z(n, d) : n \geq
0 \}$ verifies the properties of phase 1.
\begin{lemma}  \label{ell:n11}  
In the evolution of the neuronal recurrence equation $\{ z(n, d) : n
\geq 0 \}$, $\forall \ t \ \in \mathbb{N} \ {\rm such \ that} \ 0
\leq t \leq  L_1(d)+h-1-\rho(m)$, \ we \ have:
\[
z(t, d) = y(t).
\]
\end{lemma}

{\bf Phase 2.} We want that $\forall \ t \in \mathbb{N}$ such that $L_1(d)+h-\rho(m) \leq t \leq L_1(d)+h-\rho(m)+d$, we have:
\[
z(t, d) =  0 \ \ {\rm even \ when} \ y(t) = 1.
\]
Phase 2 occurs from time $h-\rho(m)+L_1(d)$ to time $L_1(d)+h-\rho(m)+d$. In the Lemma \ref{ell:n12}, we
prove that the sequence $\{ z(n, d) : n \geq 0 \}$ verifies the properties of phase 2.
\begin{lemma}   \label{ell:n12}
$\forall t \in \mathbb{N}$ {\rm \ such \ that} \ $L_1(d)+h-\rho(m) \leq t \leq L_1(d)+h-\rho(m)+d$, we have:
\[
z(t, d) \ = \ 0 \ \ {\rm even \ when} \ \ y(t) \ = \ 1.
\]
\end{lemma}

After phase 2, the behavior of the sequence $\{ z(n, d) : n \geq
0 \}$ begins to be different from the behavior of the
sequence $\{ y(n) : n \geq 0 \}.$ After phase 2, the sequence $\{ z(n, d) : n \geq
0 \}$ begins its convergence to the sequence $\{ w(n, d) : n \geq 0 \}.$  \\
{\bf Phase 3.} This phase starts at time $L_1(d)+h-\rho(m)+d+1$, and finishes at
time $L_1(d)+h-1$.
\begin{lemma}  \label{ell:n88}
In the evolution of the neuronal recurrence equation $\{ z(n, d): n
\geq 0 \}$, $\forall \ t \ \in \mathbb{N} \ {\rm such \ that} \ L_1(d)+h-\rho(m)+d+1 \leq t
 \leq L_1(d)+h-1$ \ we \ have:
\[
z(t, d) = y(t).
\]
\end{lemma}

\begin{remark} \label{rnr:nn44n} 
Based on the Lemma \ref{ell:n6}, Lemma \ref{ell:n11}, Lemma \ref{ell:n12} and Lemma \ref{ell:n88}, we easily deduce
 that the terms of the sequence $\{ z(n, d) : n \geq 0 \}$ verify the following relations:       
\begin{multline} \notag
z(tim(i,0,L_1(d)), d) z(tim(i,1,L_1(d)), d) \dots  z(tim(i,k-1,L_1(d)), d) = \\ v^{\alpha_i}(0) v^{\alpha_i}(1) \dots v^{\alpha_i}(k-1) \ , \  0 \leq \ i \ \leq d 
\end{multline}
\begin{multline}  \notag
z(tim(i,0,L_1(d)), d) z(tim(i,1,L_1(d)), d) \dots z(tim(i,k-1,L_1(d)), d) = \\ x^{\alpha_i}(1+\gamma_i(d)) x^{\alpha_i}(2+\gamma_i(d)) \dots x^{\alpha_i}(k+\gamma_i(d)) \ , \  d+1 \leq \ i \ \leq -1+\rho(m)  
\end{multline}
where $\gamma_i(d)$ is defined in Equation (\ref{eqn11c0}). 
\end{remark}

\begin{remark} 
{\rm If} 
\[
 d = \rho(m)-1
\]
{\rm Then}
it is clear that the phase 3 doesn't exists. \\
\end{remark}

{\bf Phase 4.} This phase starts at time $L_1(d)+h$ and finishes at
time $L_1(d)+h+d+\bigl(\rho(m) \times (k-1-p_{d}) \bigl).$ This phase corresponds
to the transient phase of the neuronal recurrence equation $\{ w(n, d): n \geq 0 \}$.
\begin{lemma}  \label{ell:n13}  
In the evolution of the neuronal recurrence equation $\{ z(n, d): n
\geq 0 \}$, $\forall \ t \ \in \mathbb{N} \ {\rm such \ that} \ 0 \leq t
 \leq \biggl(\rho(m) \times (k-1-p_{d}) \biggl)+d$, \ we \ have:
\[
z(L_1(d)+h+t, d) = w(h+t, d). 
\]
\end{lemma}
 
\begin{notation}
We set $L_3(d) \ = L_1(d)+2h+d - \bigl( \rho(m) \times (1+p_{d}) \bigl)$, which represents the end of the fourth phase 
and we set $L_4(d) = L_3(d)-h+1$ which represents the beginning of the cyclic phase.
\end{notation}
\begin{remark} \label{rrur:uu5} 
Based on the Lemma \ref{ell:n11}, Lemma \ref{ell:n12}, Lemma \ref{ell:n88}, 
and Lemma \ref{ell:n13}, we easily deduce that the terms of the sequence $\{ z(n, d) : n \geq 0 \}$ verify the
following relations:
\begin{multline} \notag
z(tim(i,0,L_4(d)), d) z(tim(i,1,L_4(d)), d) \dots  z(tim(i,k-1,L_4(d)), d) = \\ x^{\alpha_{i+d+1}}(\gamma_{i+d+1}(d)+k-p_d) x^{\alpha_{i+d+1}}(1+\gamma_{i+d+1}(d)+k-p_d) \dots \\ x^{\alpha_{i+d+1}}(2k-1+\gamma_{i+d+1}(d)-p_d) \ ; \ 0 \leq \ i \ \leq \rho(m)-d-2
\end{multline}
\begin{multline}
z(tim(i,0,L_4(d)), d) z(tim(i,1,L_4(d)), d) \dots z(tim(i,k-1,L_4(d)), d) = \\ 
          \underbrace{000 \dots 000}_{k} \ ; \ \rho(m)-d-1 \leq \ i \ \leq \rho(m)-1.
\end{multline}
\end{remark}
The sequence $\{ z(n, d) : n \geq \ 0 \}$ describes a cycle during
its fifth step. The subparagraph below is devoted to this study.

\subsection{Cyclic Phase}

{\bf Phase 5.} This phase starts at time $L_4(d)$ and describes a cycle of length $L_0(d)$. \\
\begin{lemma}  \label{ell:n14}
\[
z(t+L_4(d), d) \ = \ w\biggl( t+h+d+1+\Bigl(\rho(m) \times (-1-p_{d}) \Bigl), d \biggl)  \ ; \ \forall \ t \in \mathbf{N}.
\]
\end{lemma}

We have shown the following Lemma:
\begin{lemma}  \label{ecc:n1}
The sequence $\{ z(n,d) : n \geq 0 \}$ describes a transient of
length $L_4(d)$ and a cycle of length $L_0(d)$.
\end{lemma}

It is easy to see that:
\begin{itemize}
\item we can build by perturbation the neuronal recurrence equation $\{ z(n, 0) : n \geq 0 \}$ from the neuronal recurrence equation  $\{ y(n) : n \geq 0 \}$  
\item we can build by perturbation the neuronal recurrence equation $\{ z(n, d+1) : n \geq 0 \}$ from the neuronal recurrence equation $\{ z(n, d) : n \geq 0 \}$  
\end{itemize}
This second item is obtained by the following transformations:
\begin{equation}   \label{qeqq:n200}
      c(f, 1+d) =
      \begin{cases}
  c(f, d)                     ,   &\text{if $f \notin  A(d) \cup  A(1+d)$ } \\
  c(f, d)-\beta(d)+\beta(1+d) ,   &\text{if $f \in  A(d) \cap  A(1+d)$ } \\
  c(f, d)-\beta(d)            ,   &\text{if $f \in  A(d) \cap  \overline{A(1+d)}$ } \\
  c(f, d)+\beta(1+d)          ,   &\text{if $f \in  \overline{A(d)} \cap A(1+d)$ } \\
      \end{cases}
\end{equation}
\begin{equation}   \label{qeqpq:n825}
\theta_2(d+1) = \theta_2(d)  - \xi(d) + \xi(d+1).
\end{equation}
The main result of the paper is:
\begin{theorem}
$\forall m, d \in \mathbb{N}$ such that $\rho(m) \geq 2$ and $0 \leq d \leq -1+\rho(m)$. We construct a set of $\rho(m)+1$ neuronal recurrence equations which verify: 
\begin{itemize}
\item the neuronal recurrence equation $\{ y(n) : n \geq 0 \}$ describes a cycle of length $L_1(-1+\rho(m))$ 
\item by perturbation, we can build the neuronal recurrence equation $\{ z(n, 0) : n \geq 0 \}$ from the neuronal recurrence equation $\{ y(n) : n \geq 0 \}$. The period of the neuronal recurrence equation $\{ z(n, 0) : n \geq 0 \}$ is  a divisor of the period of the neuronal recurrence equation $\{ y(n) : n \geq 0 \}$
\item by perturbation, we can build the neuronal recurrence equation $\{ z(n, d+1) : n \geq 0 \}$ from the neuronal recurrence equation $\{ z(n, d) : n \geq 0 \}$. The period of the neuronal recurrence equation $\{ z(n, d+1) : n \geq 0 \}$ is  a divisor of the period of the neuronal recurrence equation $\{ z(n, d) : n \geq 0 \}$
\item the period of the neuronal recurrence equation $\{ z(n, -1+\rho(m)) : n \geq 0 \}$ is 1 (i.e. a fixed point). 
\end{itemize}
\end{theorem}
$\blacksquare$  \\

\begin{remark} \label{rrur:vv9} 
The new contribution in this paper with respect to the previous work \cite{NT 03} is that:
\begin{itemize}
\item  in the paper \cite{NT 03}, the sequence $\{ z(n)   \ : \ n  \geq 0 \} $ is a composition of the $s+1$ subsequences of periods $p_0, p_1,  \dots, p_{s}$ and $3m-1$, 
 in the evolution of the sequence  $\{ z(n)   \ : \ n  \geq 0 \} $, the subsequence of period $3m-1$ vanishes and converges to the null sequence $000  \dots 000  \dots$. 
This fact appears in the formula of transient length and in the formula of the cycle length of  the sequence $\{ z(n)   \ : \ n  \geq 0 \} $ which are respectively $(s+1)(3m+1+lcm(p_0, p_1, \dots, p_{s-1}, 3m-1 ))$ and  $(s+1) lcm(p_0, p_1, \dots, p_{s-1}, 3m-1 )$.  It is clear that in this case, the periods $p_0, p_1, \dots, p_{s-1}$ of the subsequences  intervene in 
the formula of transient and in the formula of period.
\item  in the following paper, the sequence $\{ z(n, d)   \ : \ n  \geq 0 \} $ is a composition of the $\rho(m)$ subsequences of periods $p_0, p_1,  \dots, p_{-1+\rho(m)}$ 
and in the evolution of the sequence  $\{ z(n, d)   \ : \ n  \geq 0 \} $, the subsequences of period $p_0, p_1, \dots , p_d$ vanish and converge to the null sequence $000  \dots 000  \dots$. This fact appears in the formula of transient length and on the cycle length of  the sequence $\{ z(n, d)   \ : \ n  \geq 0 \} $ which are respectively
 $\rho(m) \times lcm(p_0, p_1, \dots, p_{d})+h+d+1-\Bigl( \rho(m) \times (1+p_d) \Bigl)$, $0 \leq d \leq -1+\rho(m)$ and $\rho(m) \times lcm(p_{d+1}, p_{d+2}, \dots, p_{-1+\rho(m)})$ if $0 \leq d \leq -2+\rho(m)$ and 1 if $d=\rho(m)-1$. It is clear that in this case, there doesn't exists a period $p_i$ of a subsequence which intervenes in 
the formulas of transient and in the formula of period.
\item  the  difference mentioned in the expressions of the formula of transient and in the formula of period on the following paper and of the paper  \cite{NT 03}  imply that
 the concept used in the construction of the set $A$ ( see page 520 of the paper \cite{NT 03} )  is fondamentally different from the concept used in the construction of the set
$A(d)$ (see Equation (\ref{nre:n2d2})).
\item in the paper \cite{NT 03}, we build only one neuronal recurrence equation $\{ z(n)   \ : \ n  \geq 0 \} $ whereas on the following paper we build $ \rho(m)$ neuronal recurrence
 equation $\{ z(n, d)   \ : \ n  \geq 0 \} $, $0  \leq d  \leq   \rho(m)-1$.
\end{itemize}
\end{remark}

Let us note $e$ an integer such that:
\[
\forall \ i \ \in \mathbb{N} \ , \ 0 \leq i \leq \rho(m)-1 \ {\rm we \ have \ } \ \beta(m, \alpha_e) \leq \beta(m, \alpha_i)
\]
Subsequently, we suppose that:
\[
d \ < \ \beta(m, \alpha_e).
\]

Let us note $\widetilde{z}(0,d) \widetilde{z}(1, d) \dots \widetilde{z}(h-1, d)$ the following $h$ terms:
\begin{align}
&\widetilde{z}(i, d) \ = \ y(i) \ \ {\rm for} \ i \ {\rm such \ that \ } \beta(m,\alpha_e)-d \leq i \leq h-1   \\
&\widetilde{z}(i, d)  \ \in \{ 0, 1 \} \ {\rm for} \ i \ {\rm such \ that \ } 0 \leq i \leq \beta(m,\alpha_e)-d-1 
\end{align}
The following lemma characterizes the basin of attraction of the sequence $\{ z(n, d) \ : \ n \geq 0 \}$.
\begin{lemma}  \label{ch29:eq4}
{\bf If}  $d \ < \ \beta(m, \alpha_e)$  {\bf Then} from the following initial configurations:
\begin{align}   \notag
 &y(0)y(1) \dots y(h-1) ,       \notag  \\
 &\widetilde{z}(0, d) \widetilde{z}(1, d) \dots \widetilde{z}(h-1, d)  \notag
\end{align}
the neuronal recurrence equation $\{ z(n, d) \ : \ n \geq 0 \}$ converges to the same basin of attraction. 
\end{lemma}

\section{Conclusion}  We  improve the fundamental lemma of composition of neuronal recurrence equations.  From a neuronal recurrence equation that describes a cycle of length $L_2$, we construct a set of $\rho(m)$ neuronal recurrence equations $\{z(n,d) : n \geq 0 \}$ whose dynamics describe respectively the cycle of length $L_0(d)$ where $L_0(d)$ are the divisors of $L_2$. The neuronal recurrence equation $\{z(n,-1+\rho(m)) : n \geq 0 \}$ describes an exponential transient and a fixed point. By perturbation, we have built the neuronal recurrence equation $\{z(n,d+1) : n \geq 0 \}$  from the neuronal recurrence equation $\{z(n,d) : n \geq 0 \}$ such that the period of the neuronal recurrence
equation $\{z(n,d+1) : n \geq 0 \}$ is a divisor of the period of the neuronal recurrence equation 
$\{z(n,d) : n \geq 0 \}$. Thus, we have built a period-halving bifurcation of a neuronal recurrence equation. This result is inscribed in the framework of results on convergence time of  neural  networks \cite{GM 89, SO 00, SO 20}. The exponential convergence time of neuronal recurrence equations can be useful when we want to use it in cryptographic toolbox ( e.g.  remote  authentication,  generation of pseudo-random numbers, ... ).     \\

{\bf  Acknowledgements}   \\
The author express his gratitude to  Professor  Michael   Y.  Li  ( University of  Alberta )  for  useful  discussions.  The author also express his gratitude 
to the reviewer for useful remarks.  This work was supported by   {\it  LIRIMA } ,   {\it University of Yaounde 1 }, and  by   {\it The Abdus Salam International Centre for Theoretical Physics (ICTP)}, 
Trieste, Italy.

\appendix{\bf Appendix}   \\
{\bf Proof of Lemma  \ref{elll:ss8} } \\ 
The proof is subdivided into  two cases:  \\
{\it First case:} $\exists \ j, \ 1 \leq \ j  \ \leq \ r$ such that $p_j \geq 2$ \\
This case corresponds to Lemma \ref{elll:nn1} and it was proved by Cosnard, Tchuente and Tindo \cite{CTT 92}. \\
{\it Second case:}   $\forall \ j, \ 1 \leq \ j  \ \leq \ r$ we have $p_j = 1$. \\
Without loss of generality, we suppose that the transient length of each sequence is $0$. It is easy to see that:
\[
p_j \ = \ 1 \ \Longrightarrow \ x^j(n) = x^j(0) \ , \ \forall \ n \in \mathbb{N}.
\]
It follows that $Per$ is the period of the following sequence:
\begin{multline}
x^1(0) x^2(0) \dots x^r(0) x^1(1) x^2(1) \dots x^r(1) \cdots x^1(\ell) x^2(\ell) \dots x^r(\ell) \dots  =  \\ 
   x^1(0) x^2(0) \dots x^r(0) x^1(0) x^2(0) \dots x^r(0) \cdots x^1(0) x^2(0) \dots x^r(0) \dots
\end{multline}
It is clear that $r$ is a multiple of the period, i.e. $Per$ is a divisor of $r$. \\
$\blacksquare$   \\

{\bf Proof of Lemma  \ref{elnll:puon2} } \\ 
 Without loss of generality, we can assume that for all $\jmath, \ 1 \leq \jmath < g$, we have $T_{\jmath} \leq T_g$. It suffices to apply the construction used in the demonstration of Lemma \ref{elll:nn1} to the sequences $\{ x^{\jmath}(n) : n \geq 0 \} \ , 1 \leq \jmath \leq g$.  \\
$\blacksquare$  \\ 

{\bf Proof of Proposition  \ref{ch9:prop1} } \\ 
 The proof is given in \cite{NT 04} (see Proposition 1 of \cite{NT 04}). \\
$\blacksquare$  \\

{\bf Proof of Proposition \ref{ch9:prop22}   } \\ 
 The proof is subdivided into  two cases:  \\
{\bf First case:}   $\rho(m)$ is even  \\
\begin{align}
\sum_{j \in Pos(\alpha_i)}  \bar{a}_j  &=   \sum_{\ell=1}^{2 \times \rho(m)} \bar{a}_{\ell \times p_i}   \\
                                       &=   \sum_{\ell=1}^{(3 \times \rho(m))/2} 2 - \sum_{\ell=1+((3 \times \rho(m))/2) }^{2 \times \rho(m)} 2     \\
                                       &= 2 \times \rho(m)
\end{align}

{\bf Second case :}   $\rho(m)$ is odd and $\rho(m) \geq 3$ \\

\begin{align}
\sum_{j \in Pos(\alpha_i)}  \bar{a}_j  &=   \sum_{\ell=1}^{2 \times \rho(m)} \bar{a}_{\ell \times p_i}   \\
                                       &=   \sum_{\ell=1}^{((3 \times \rho(m))-1)/2} 2 - \sum_{\ell=((3 \times \rho(m))+1)/2}^{(2 \times \rho(m))-2} 2 - \sum_{\ell=(2 \times \rho(m))-1}^{2 \times \rho(m)} 1   \\
                                       &= 2 \times \rho(m)
\end{align}
$\blacksquare$  \\

{\bf Proof of Lemma  \ref{ch9:lem1} } \\ 

We have:
\begin{align}  \notag
\sum_{j=1}^{k} \bar{a}_{j} x^{\alpha_i}(k-j) & = \sum_{j \in Pos(\alpha_i)} \bar{a}_{j} 
   x^{\alpha_i}(k-j) +  \sum_{j \in D \setminus Pos(\alpha_i)} \bar{a}_{j} 
   x^{\alpha_i}(k-j)   \notag  \\
   & =  \sum_{j \in Pos(\alpha_i)} \bar{a}_{j} x^{\alpha_i}(k-j) \ {\rm \ by \ application \ of 
                          \ the \ Equation (\ref{ch9:eq8}) }  \notag  \\
   & =  \sum_{j \in Pos(\alpha_i)} \bar{a}_{j} \ \ \ {\rm \ by \ application \ of \ the \
                            \ Equation (\ref{ch9:eq7}) }  \notag    \\
   & =  2 \times \rho(m) \ {\rm \ by \ application \ of \ the \ Proposition \ \ref{ch9:prop22} }   \label{equat:22}
\end{align}
It follows that:
\begin{align}  \notag
x^{\alpha_i}(k)  & = {\bf 1} \Bigl( \sum_{j=1}^{k} \bar{a}_{j}  x^{\alpha_i}(k-j) -
                                \bar{\theta}  \Bigl)   \notag   \\
   & =  {\bf 1} \Bigl( \bar{\theta}  -  \bar{\theta}  \Bigl) \ \ \ 
                     {\rm \ by \ application \ of \ the \ Equation \ (\ref{equat:22})} \notag \\
   & =   1   \notag  
\end{align}
$\blacksquare$ \\

{\bf Proof of Lemma  \ref{ch9:lem2} } \\ 
The result follows from the definition of $E(\alpha_i, d)$
and by application of Proposition \ref{ch9:prop1}. \\
$\blacksquare$ \\

{\bf Proof of Lemma  \ref{ch9:lem3} } \\ 
By application of Lemma \ref{ch9:lem1} and Lemma
\ref{ch9:lem2} we deduce the result. \\
$\blacksquare$  \\

{\bf Proof of Lemma  \ref{ellll:n3} } \\ 
From Equation (\ref{ch9:eq4}) and Equation (\ref{ch9:eq5}) we deduce the result. \\
$\blacksquare$ \\

{\bf Proof of Lemma  \ref{ell:n4} } \\ 
From Lemma \ref{ellll:n3}, we have $\sum_{j=1}^{k}
x^{\alpha_i}(n-j) \ \leq \ 1+\mu(m, \alpha_i)$, therefore it follows that $-2(1+\mu(m, \alpha_i)) \ \leq \
S1(\alpha_i, n)$ because $-2 \leq \bar{a}_j$ for $j$ such that $1 \leq j \leq k$. Since $\bar{a}_j$ is an integer, on the basis of Equation (\ref{eeq:n1}), Equation (\ref{eeeqq:n2}) and Equation (\ref{eeq:n333}), we easily deduce that
\[
S1(\alpha_i,n) \  < \ \bar{\theta}  \Longrightarrow \ \ S1(\alpha_i, n) \leq \bar{\theta} - 1.
\]
From the evolution of the sequence $\{ x^{\alpha_i}(n) : n \geq 0 \}$,
we know that for $t \ \in \{k, k+1, ..., k+3m-\alpha_i-2, k+3m-\alpha_i-1 \}$ the
only $t$ at which $x^{\alpha_i}(t) = 1$ is $k$ and that $S1(\alpha_i, k)=
\bar{\theta}$. From the fact that the sequence $\{ x^{\alpha_i}(n) : n \geq 0
\}$ has period $3m-\alpha_i$, we deduce that
\[
S1(\alpha_i,n) \  \geq \ \bar{\theta}  \Longrightarrow \ \ S1(\alpha_i, n) \ = \bar{\theta}.
\]
From the fact that $\forall \ n \ \in  \mathbb{N}$,
\[
S1(\alpha_i, n) \ \in \ [ -2(1+\mu(m, \alpha_i)) , \bar{\theta} - 1 ]   \cup  \{ \bar{\theta} \}.
\]
It follows from the definition of $\lambda$ that $\lambda \ \in [ -1, 0 [$. \\
$\blacksquare$ \\

{\bf Proof of Lemma  \ref{ell:n5} } \\ 
From Equation (\ref{eeq:n7}), we have:
\[
v^{\alpha_i}(k-1) \ = \ \overline{x^{\alpha_i}(k)} \ = \ 0.
\]
From Equations (\ref{ch9:eq4}), (\ref{ch9:eq5}) and (\ref{eeq:n7}), we easily deduce that the structure of the
initial terms of the sequence $\{ v^{\alpha_i}(n) : n \ \geq \ 0 \}$ have the following form:
\begin{equation}  \label{c9h9:e5qq5}
v^{\alpha_i}(0) v^{\alpha_i}(1) \dots v^{\alpha_i}(k-1) = \underbrace{0 \dots 0}_{-1+\beta(m, \alpha_i)} 
\underbrace{\underbrace{10 \dots 0}_{p_i} \cdots \underbrace{10 \dots 0}_{p_i}}_{\mu(m, \alpha_i)p_i }0
\end{equation}
We want to prove by recurrence the part (a) of the Lemma: \\
{\bf Basis Case:} \ $t = k$   \\
Based on Equations (\ref{eeeqq:n2}), (\ref{eeq:n333}) and (\ref{c9h9:e5qq5}), we easily deduce that:
\begin{equation}   \label{c9h9:e6qq6}
\sum_{j=1}^{k} \bar{a}_j v^{\alpha_i}(k-j) \ \leq \ \bar{\theta}-2.
\end{equation}
This implies that $v^{\alpha_i}(k) \ = \ 0$.  \\
{\bf Recurrence Hypothesis:} we suppose that the property is true at all steps $\ell-k, \ell-k+1, \cdots, \ell-1$, i.e.
\begin{equation} \label{c9h9:e8qq7}
\forall \ t \in \mathbb{N} {\rm \ such \ that} \ k \leq t \leq \ell-1, \ {\rm we \ have} \ v^{\alpha_i}(t) \ = \ 0
\end{equation}
and \\
$\forall \ t \in \mathbb{N} {\rm \ such \ that: \ } t \leq k-1$, $v^{\alpha_i}(t)$ is defined by Equation (\ref{c9h9:e5qq5}). \\
{\bf Step $\ell$:}  \\
Based on Equation (\ref{c9h9:e8qq7}), we easily deduce that:
\begin{equation} \label{c9h9:e9qq8}
\sum_{j=1}^{k} \bar{a}_j v^{\alpha_i}(\ell-j) \ \leq \ \bar{\theta}-2.
\end{equation}
This implies that $v^{\alpha_i}(\ell) = 0$. We have shown the part (a) of the Lemma.  \\
The proof of part (b) is easily deduced from Equations (\ref{c9h9:e6qq6}) and (\ref{c9h9:e9qq8}).  \\
The part (c) is easily deduced from the part (a).  \\
$\blacksquare$ \\

{\bf Proof of Lemma  \ref{ell:n7} } \\ 
By applying the Lemma \ref{elll:nn1} and given the fact that
the subsequences $\{ x^{\alpha_i}(n) : n \geq 0 \}$ describe a cycle
of length $p_i$, we deduce the result.  \\ $\blacksquare$ \\

{\bf Proof of Lemma  \ref{ell:n9} } \\ 
By application of Lemma \ref{elnll:puon2} and given the fact that:
\begin{itemize}
 \item the subsequence $\{ v^{\alpha_i}(n) \ : \ n \geq 0  \}$ describes a transient of length $k-p_i$ and the following fixed point
$\underbrace{000 \dots 0000}_k$,
 \item $p_{-1+\rho(m)} \ < \ \dots \ < \ p_{i+1} \ < \ p_i \ < \dots < \ p_1 \ < \ p_0$,
\item the subsequence $\{ x^{\alpha_i}(n) \ : \ n \geq 0  \}$ describes a cycle of length  $p_i$,
\end{itemize}
we deduce the result.
\\ $\blacksquare$  \\

{\bf Proof of Lemma  \ref{ell:n8} } \\ 
Based on the fact that the sequence $\{ y(n) : n \geq 0 \}$ is the shuffle of $\rho(m)$ subsequences of the form $\{x^{\alpha_i}(n) : n \geq 0 \}$ and on
the Lemma \ref{ellll:n3}, we deduce that:
\[
\sum_{i=0}^{-1+\rho(m)} \mu(m, \alpha_i) \ \leq \sum_{f=1}^{h} y(n-f) \ \leq \rho(m) + \sum_{i=0}^{-1+\rho(m)} \mu(m, \alpha_i).
\]
From the facts that:
\begin{itemize}
 \item $\sum_{f=1}^{h}  v^{\alpha_i}(h-f) \ \leq \ \mu(m, \alpha_i)$
 \item $\forall n \geq h \ , \ v^{\alpha_i}(n) = 0$
 \item $\{ w(n, d) : n \geq 0 \}$ is the shuffle of $d+1$ subsequences of the form $\{ v^{\alpha_i}(n) : n \geq 0 \}$
 and of $\rho(m)-d-1$ subsequences of the form $\{ x^{\alpha_i}(n) : n \geq 0 \}$
\end{itemize}
we deduce that:
\[
\sum_{i=d+1}^{-1+\rho(m)} \mu(m, \alpha_i) \ \leq \sum_{f=1}^{h} w(n-f, d) \ \leq -d-1+\rho(m)+\sum_{i=0}^{-1+\rho(m)} \mu(m, \alpha_i).
\]
$\blacksquare$   \\

{\bf Proof of Proposition \ref{epp:n1} } \\ 
The state of the neuron $z$ at time $n$ is given by the formula:
\begin{flalign}   
z(n, d) &= {\bf 1} \left[ \sum_{f=1}^{h} c(f, d)z(n-f, d)- \theta_2(d) \right]   \notag   \\
     &= {\bf 1} \left[ \sum_{1 \leq f \leq h, \ f \notin A(d)} c(f,d)z(n-f,d) +
       \sum_{ f \in A(d) } c(f,d)z(n-f,d) - \theta_1 - \xi(d) \right]   \notag   \\
     &={\bf 1} \left[ \sum_{1 \leq f \leq h, \ f \notin A(d)} b_{f}z(n-f,d) +
       \sum_{ f \in A(d) } b_{f}z(n-f,d) + Q2(n, d) - \theta_1  \right]   \label{eeq:n26} 
\end{flalign}

By hypothesis, we known that $n-f \equiv 0 \pmod{\rho(m)} \ {\rm implies \ that } \ z(n-f,d) \ = \ y(n-f)$. From Equation (\ref{eeq:n11}), we have $f \not\equiv 0 \pmod{\rho(m)}$ implies that $b_f = 0$. Consequently, it follows that:

\begin{flalign}
z(n, d) &= {\bf 1} \left[\sum_{1 \leq f \leq h, \ f \equiv 0 \pmod{\rho(m)} } b_{f}y(n-f) + Q2(n, d) - \theta_1 \right]   \notag   \\
     &= {\bf 1} \left[ S2(n)+ Q2(n, d)- \theta_1 \right]   \notag
\end{flalign}
$\blacksquare$ \\

{\bf Proof of Proposition \ref{epp:n2} } \\ 
By hypothesis, we have $\sum_{f \in A(d)} \ z(n-f, d) \ = \ Tot(d)$. It
follows that:
\begin{align} \notag
\beta(d) \ \sum_{f \in A(d)} \ z(n-f, d) \ &= \ \beta(d) \times Tot(d) \notag \\
                                  &= \ \lambda.   \notag
\end{align}
We deduce that: $Q2(n, d) \ = \ \frac{\beta(d)}{8}$. \\
$\blacksquare$ \\

{\bf Proof of Proposition \ref{epp:n3} } \\ 
By hypothesis, we have $\sum_{f \in A(d)} z(n-f, d) \ \leq
\ Tot(d)-1$. It follows that
\begin{align} \notag
   \beta(d) \times (Tot(d) -1)  & \leq  \beta(d) \sum_{f \in A(d)} z(n-f, d) \notag \\
   \beta(d) \times (Tot(d) -1) - \xi(d)  & \leq  \beta(d) \sum_{f \in A(d)} z(n-f, d) - \xi(d).  \notag
\end{align}
We obtain:
\[
 \frac{-7 \beta(d)}{8} \leq Q2(n, d).
\]
Likewise, we have :  $0 \leq \sum_{f \in A(d)} z(n-f, d)$. It follows
that :
\begin{align}   \notag
\beta(d) \ \sum_{f \in A(d)} \ z(n-f, d) \ \  & \leq  0    \notag  \\
\beta(d) \ \sum_{f \in A(d)} \ z(n-f, d) - \xi(d) \ \ & \leq  -\xi(d) \notag  \\
Q2(n, d)  &\leq  - \xi(d).       \notag
\end{align}
We deduce that: $\frac{-7 \beta(d)}{8} \ \leq Q2(n, d) \ \leq \ - \xi(d)$. \
\\ $\blacksquare$  \\

{\bf Proof of Lemma \ref{lem:mel88} } \\ 
The proof is subdivided into  two parts:  \\
{\it First part:}  \\ Based on the facts that:
\begin{itemize}
\item $0 \leq d \leq -1+\rho(m) \ \leq 2m \ \leq p_i \ ,$ for i such that $0 \leq i \leq -1+\rho(m)$
\item From Equation (\ref{zzr:rzz2}), we deduce that $\rho(m) \geq 2$ implies that $\mu(m, \alpha_i) \geq 4$ 
\item the fact that $\{ y(n) \ : \ n \geq 0 \}$ is the shuffle of $\rho(m)$ subsequences $\{ x^{\alpha_i}(n) \ : \ n \geq 0 \}$ 
\item the structure of the subsequence $\{ x^{\alpha_i}(n) \ : \ n \geq 0 \}$ defined in Equation (\ref{ch9:eq4})
\end{itemize}
we easily deduce that it is possible by using only $B_0(d), p_0, p_1, \dots, p_{-1+\rho(m)}$ to construct all the terms
\[
y(j) \ {\rm \ such \ that \ } \ 1-\rho(m)+L_1(d) \leq j \leq -\rho(m)+h+L_1(d).
\]
{\it Second part:}  \\ The proof will be done by contradiction. Let us suppose that :
\begin{align}
&n \not\equiv h-\rho(m)+L_1(d) \pmod{L_2}      \label{pop:opp8}   \\
&\sum_{f \in B_0(d)} y(n-f)  \ = \ Tot(d)       \label{pop:opp6}  
\end{align}
Based on the First part, on Equation (\ref{ren:nd4}), on the period of the sequence $\{ y(n) \ : \ n \geq 0 \}$, we deduce that
\[
 n \equiv h-\rho(m)+L_1(d) \pmod{L_2}      
\]
This implies a contradiction with Equation (\ref{pop:opp8}). \\
$\blacksquare$  \\

{\bf Proof of Lemma \ref{lem:mel99} } \\ 
Based on:
\begin{itemize}
\item the definition of the set $A(d)$ 
\item the construction of the set $A(d)$ from $B_i(d) \ ; 0 \leq i \leq d$
\item the Lemma \ref{lem:mel88}
\end{itemize}
we deduce the result. \\
$\blacksquare$  \\

{\bf Proof of Lemma \ref{ell:n11} } \\      
The proof will be done by induction on $t$: \\
{\bf Basic Case:} steps $0 \leq t \leq h - 1$   \\
The result can be deduced from Equation (\ref{eeq:n22}). \\
{\bf Induction hypothesis:} we suppose the property is true
at all steps $n-1$ such that $h-1 \leq n-1 \leq L_1(d)+h-2-\rho(m).$ \\
{\bf Step n.} \\
Firstly, we want to show that $\sum_{f \in A(d)} y(n-f) \ \leq Tot(d)-1$.  \\
From induction hypothesis, we easily deduce that: $n \leq L_1(d)+h-1-\rho(m)$. By application of Lemma \ref{lem:mel99}, 
 we conclude that: 
\[
 \sum_{f \in A(d)} y(n-f) \leq Tot(d)-1.
\]
Consequently, by applying Proposition \ref{epp:n3}, we deduce that 
\begin{equation} \label{eeq:n27}
\frac{-7 \beta(d)}{8}  \leq  Q2(n, d) \leq  -\xi(d)
\end{equation}
Secondly, we want to show that $z(n,d) \ = \ y(n)$.
From the induction hypothesis and Proposition \ref{epp:n1}, we
know that: $z(n,d) = {\bf 1} \left[ S2(n)+Q2(n,d)- \theta_1 \right]$. Given
the value of $S2(n)$, we shall distinguish two cases: \\
{\bf Case 1 :}    $S2(n) \ \geq \ \theta_1$ \\
This implies that $y(n) = 1$. From Equation (\ref{eeq:n27}), 
we have $0 < Q2(n)$. It follows that: 
\[
0 \ < \ \frac{-7 \beta(d)}{8}  \leq  S2(n)+Q2(n, d)- \theta_1.
\]
Consequently: $z(n,d) = 1 = y(n).$ \\
{\bf Case 2 :}  $S2(n) \ < \ \theta_1$  \\
This implies that $y(n) = 0$. From Remark \ref{r:n2}, we have 
$S2(n) - \theta_1 \ \leq \lambda$. From
Equation (\ref{eeq:n27}), we have $Q2(n,d) \ \leq -\xi(d)$. It follows
that $S2(n) + Q2(n,d) - \theta_1 \ \leq \lambda -\xi(d) < \
0$. Consequently $z(n,d) = 0 = y(n)$.  \\
$\blacksquare$ \\

{\bf Proof of Lemma \ref{ell:n12} } \\    
From Equation (\ref{eeq:n9}), Lemma \ref{ch9:lem1} and Lemma \ref{ch9:lem2}, it is easy to deduce that:
\begin{equation} \label{rer:rer4rer}
 y(t) = 1 \ {\rm i.e.} \ S2(t) = \theta_1 \ \ {\rm for \ t \ such \ that \ } h-\rho(m) \leq t \leq h-\rho(m)+d.
\end{equation}
Based on the periods of the subsequences $\{ x^{\alpha_i}(n) : n \geq 0 \}$, we conclude that:
\begin{equation} \label{rer:rer5rer}
 y(t) = 1 \ {\rm i.e.} \ S2(t) = \theta_1 \ \ {\rm for \ t \ such \ that \ }  L_1(d)+h-\rho(m) \leq t \leq L_1(d)+h-\rho(m)+d
\end{equation}
From:
\begin{itemize}
 \item the definition of the sets $B_i(d) \ (0 \leq i \leq d)$ and $A(d)$
 \item Proposition \ref{epp:n2}
\end{itemize}
we conclude that:
\begin{equation} \label{rer:rer7rer}
Q2(t, d) = \frac{\beta(d)}{8} \ \ {\rm for \ t \ such \ that \ }  L_1(d)+h-\rho(m) \leq t \leq L_1(d)+h-\rho(m)+d
\end{equation}
From Proposition \ref{epp:n1} and Lemma \ref{ell:n11}, we deduce that:
\begin{equation} \label{rer:rer9rer}
z(t, d) = {\bf 1} \left[ S2(t)+Q(t, d)- \theta_1 \right].
\end{equation}
From Equation (\ref{rer:rer5rer}), Equation (\ref{rer:rer7rer}) and Equation (\ref{rer:rer9rer}), we conclude that:
\begin{align}
 z(t, d) &= {\bf 1} \left[ \theta_1+ \frac{\beta(d)}{8} - \theta_1 \right]  \notag  \\
      &= {\bf 1} \left[ \frac{\beta(d)}{8} \right]  \notag  \\
      &= 0 \notag 
\end{align}
$\blacksquare$   \\

{\bf Proof of Lemma \ref{ell:n88} } \\    
From Proposition \ref{epp:n3}, Lemma \ref{lem:mel99}, Lemma \ref{ell:n11} and Lemma \ref{ell:n12}, we deduce that: 
\begin{equation}   \label{eqe:r44}
\frac{-7 \beta(d)}{8}  \ \leq \ Q2(t, d) \ \leq \ -\xi(d).
\end{equation}
From Proposition \ref{epp:n1}, we have:
\[ 
z(t, d) = {\bf 1}[ S2(t) + Q2(t, d)- \theta_1].
\]
Based on the value of $S2(t)$, we can distinguish two cases:  \\
{\bf First Case:}  $S2(t)= \theta_1$  \\
If follows that
\[
0 \leq \frac{-7 \beta(d)}{8}  \leq  S2(t) + Q2(t, d)- \theta_1
\]
Consequently
\[
 z(t, d) \ = \ y(t) = 1 
\]
{\bf Second Case:}  $S2(t) \ < \theta_1$ \\
Based on Remark \ref{r:n2} and Equation (\ref{eqe:r44}), it follows that:
\[
S2(t) + Q2(t, d)- \theta_1  \leq \lambda - \xi(d) \ < \ 0
\]
Consequently
\[
 z(t, d) \ = \ y(t) = 0
\]
$\blacksquare$  \\

{\bf Proof of Lemma \ref{ell:n13} } \\  
The proof will be done by induction on $t$. \\
{\bf Basic Case:}  \\
{\bf Induction hypothesis:} assuming that: 
\begin{align}
&z(h+L_1(d)-i, d) = w(h-i, d) \ {\rm for} \ i \ {\rm such \ that} \ 1 \leq i \leq h \label{pp8:p8}  \\
&z(h+L_1(d)+j, d) = w(h+j, d) \ {\rm for} \ j \ {\rm such \ that} \ 0 \leq j \leq \ell-1 \label{pp0:p2} 
\end{align}
{\bf Step $\ell$:}  \\
\begin{align}  
z(\ell+h+L_1(d), d) &= {\bf 1} \left[ \sum_{f = 1}^{h} c(f, d) z(-f+\ell+h+L_1(d), d) -
                                \theta_2(d) \right]   \label{eeq:n301}    \\
   &= {\bf 1} \left[\sum_{f = 1}^{h} b_f z(-f+\ell+h+L_1(d), d) -
                    \theta_1 + Q2(\ell+h+L_1(d), d) \right]. \label{eeq:n302}
\end{align}
From (a) of Lemma \ref{ell:n5}, Lemma \ref{ell:n11},  Lemma
\ref{ell:n12} and Lemma \ref{ell:n88} we have:
\begin{equation}  \label{eeq:n31}  
\sum_{f \ \in  \ A(d)} z(-f+\ell+h+L_1(d), d) \ \leq \ \sum_{f \ \in \ A(d)}
y(-f+\ell+h+L_1(d)).
\end{equation}
From Proposition \ref{epp:n3}, Lemma \ref{lem:mel99} and Equation (\ref{eeq:n31}) we deduce that 
\begin{equation}   \label{eeq:n33}
\frac{-7 \beta(d)}{8}  \ \leq \ Q2(\ell+h+L_1(d), d) \ \leq \ -\xi(d).
\end{equation}
From Equation (\ref{eeq:n302}), by hypothesis defined by Equation (\ref{pp8:p8}) and by hypothesis defined by Equation (\ref{pp0:p2}), we can write:
\begin{align}  \notag
z(\ell+h+L_1(d), d) &= {\bf 1} \left[ \sum_{f=1}^{h} b_{f} w(\ell+h-f, d) - \theta_1 + Q2(\ell+h+L_1(d), d) \right] \notag  \\
                    &= {\bf 1} \left[ S3(\ell+h, d) - \theta_1 + Q2(\ell+h+L_1(d), d) \right]   \notag
\end{align}
The proof will be subdivided into  two cases:  \\
{\bf First Case:} $\ell \equiv  \jmath  \pmod{\rho(m)}$ and $\jmath \in \{ 0, 1, 2, \dots, d \}$  \\
By basing ourselves on the part (a) of Lemma \ref{ell:n5}, Remark \ref{rnr:nn44n}, Lemma \ref{ell:n11}, Lemma \ref{ell:n12} and Lemma \ref{ell:n88}, we deduce that: 
\begin{equation} 
S3(\ell+h, d) -\theta_1   \leq   -2.
\end{equation}
By using Equation (\ref{eeq:n33}), we deduce that:
\[
S3(\ell+h, d) - \theta_1 + Q2(\ell+h+L_1(d), d) \leq -2-\xi(d) \ < \ 0
\]
It follows that:
\[
z(\ell+h+L_1(d), d) = 0 = w(h+\ell, d)
\]
{\bf Second Case:} $\ell \equiv  \jmath  \pmod{\rho(m)}$ and $\jmath \in \{ d+1, d+2, \dots, -1+\rho(m) \}$  \\
Let us note $\ell = (\imath \times \rho(m))+\jmath$ where $\imath \in \mathbb{N}$ and $\jmath \in \{d+1, d+2, \dots, -1+\rho(m) \}$.
Based on the structure of $b_f$ (see Equation (\ref{eeq:n11})) and on Remark \ref{rnr:nn44n}, we deduce that:
\[
S3(\ell+h, d) = \sum_{f=1}^{k}  \bar{a}_f x^{\alpha_{\jmath}}(k+1+\gamma_{\jmath}(d)+\imath-f) 
\]
Based on the value of $S3(\ell+h, d)$, we have to distinguish two subcases:  \\
{\it Subcase 1:}    $S3(\ell+h, d) = \theta_1$  \\
The hypothesis $S3(\ell+h, d) = \theta_1$ implies that $w(h+\ell, d) = 1$. From Equation (\ref{eeq:n33}) we deduce that:
\[
0 \leq \frac{-7 \times \beta(d)}{8} \leq S3(\ell+h, d) - \theta_1 + Q2(\ell+h+L_1(d), d) 
\]
It follows that:
\[
 z(\ell+h+L_1(d), d) = 1 = w(h+\ell, d) 
\]
{\it Subcase 2:}  $S3(\ell+h, d) < \theta_1$ \\
The hypothesis $S3(\ell+h, d) \ < \ \theta_1$ implies that $w(h+\ell, d) = 0$. From Remark \ref{r:n2} and the hypothesis, we deduce that:
\[
S3(\ell+h, d) - \theta_1  \leq \lambda
\]
From Equation (\ref{eeq:n33}) we deduce that:
\[
S3(\ell+h, d) - \theta_1  + Q2(\ell+h+L_1(d), d) \leq \lambda - \xi(d)  < 0
\]
It follows that:
\[
z(\ell+h+L_1(d), d) = 0 = w(h+\ell, d).
\]
$\blacksquare$ \\

{\bf Proof of Lemma \ref{ell:n14} } \\  
The proof shall be done by a strong induction on $t$.  \\
{\bf  Basic Case:} steps $0, 1, 2, \dots, h-1$  \\
With regards: 
\begin{itemize}
\item to the configuration of the sequence $\{ z(n, d) : n \geq \ 0 \}$ at the end of phase 4 (Remark \ref{rrur:uu5})
\item the transient length of the sequence $\{ w(n, d) : n \geq \ 0 \}$
\item the initial configuration of the sequence $\{ w(n, d) : n \geq \ 0 \}$ (see Equations (\ref{ComNov:n09}) and (\ref{ComNov:n19}))
\end{itemize}
we deduce that 
\[
z(f+L_4(d), d) \ = \ w\biggl( f+h+d+1+\Bigl(\rho(m) \times (-1-p_{d}) \Bigl), d \biggl) \ ; \forall \ f \in \mathbb{N}, \ 0 \leq \ f \leq h-1
\]
{\bf Induction hypothesis:} steps $\ell-h, \ell-h+1, \dots, \ell-1$ \\
We suppose that the property holds for all $\jmath$ such that
$\ell-h \ \leq \jmath \ \leq \ \ell-1$, i.e., 
\begin{equation}   \label{eeq:n412}
z(\jmath+L_4(d), d) = w\biggl( \jmath+h+d+1+\Bigl(\rho(m) \times (-1-p_{d}) \Bigl), d \biggl) 
\end{equation}
{\bf Step $\ell$:}
It is clear that
\begin{equation}  \label{mpo:mopop99}
\ell \geq h 
\end{equation}
We also have:
\begin{align} 
z(\ell+L_4(d), d) &= {\bf 1}\left[\sum_{f=1}^{h} c(f, d) z(-f+\ell+L_4(d), d) - \theta_2(d)\right]  \label{eeq:n411}  \\
&= {\bf 1} \left[ \sum_{f=1}^{h} b_f z\bigl(-f+\ell+L_4(d), d \bigl) - \theta_1 + Q2(\ell+L_4(d), d) \right] \label{eeq:n41224}  \\
&= {\bf 1} \left[ \sum_{f=1}^{h} b_f w\biggl(-f+\ell+h+d+1+\bigl(\rho(m) \times (-1-p_{d})\bigl), d \biggl) - \theta_1 + Q2(\ell+L_4(d), d) \right]  \label{eeq:n622} \\
&= {\bf 1} \left[ S3\biggl(\ell+h+d+1+\bigl(\rho(m) \times (-1-p_{d}) \bigl), d \biggl) - \theta_1 + Q2(\ell+L_4(d), d) \right]  \label{eeq:n6p29}
\end{align}
From Equation (\ref{mpo:mopop99}), we have $\ell \geq h$. By using Lemma \ref{lem:mel99}, we deduce that:
\begin{equation}   \label{aut:aut1}
\sum_{f \in A(d)} y(-f+\ell+L_4(d)) \ \leq \ Tot(d)-1 
\end{equation}
From Lemma \ref{ell:n11}, Lemma \ref{ell:n12}, Lemma \ref{ell:n88} and Lemma \ref{ell:n13}, we easily deduce that:
\begin{equation}  \label{mpo:mopop88}
\sum_{f \in A(d)} z(-f+\ell+L_4(d), d) \ \leq \ \sum_{f \in A(d)} y(-f+\ell+L_4(d))
\end{equation}
From Equations (\ref{aut:aut1}) and (\ref{mpo:mopop88}),  we easily deduce that:
\begin{equation}  \label{aut:aut3}
\sum_{f \in A(d)} z(-f+\ell+L_4(d), d) \leq Tot(d)-1
\end{equation}
Consequently by applying Proposition \ref{epp:n3}, we deduce that:
\begin{equation}  \label{p78:ppu78}
\frac{-7 \beta(d)}{8}  \ \leq \ Q2(\ell+L_4(d), d) \ \leq \ -\xi(d).
\end{equation}
Let us suppose that $\ell = (i_1 \times \rho(m)) + i_2$ with $0 \leq i_2 \leq -1+\rho(m)$. We have to distinguish two cases:  \\
{\bf First Case:}  $0 \leq i_2 \leq \rho(m)-d-2$  \\
Based on the value of $S3\biggl(\ell+h+d+1+\bigl(\rho(m) \times (-1-p_{d})\bigl), d \biggl)$, we have to distinguish two subcases: \\
{\it Subcase 1:} $S3\biggl(\ell+h+d+1+\bigl(\rho(m) \times (-1-p_{d}) \bigl), d \biggl)  = \theta_1$  \\
The hypothesis $S3\biggl(\ell+h+d+1+\bigl(\rho(m) \times (-1-p_{d}) \bigl), d \biggl) = \theta_1$ implies that $w\biggl(\ell+h+d+1+\bigl(\rho(m) \times (-1-p_{d}) \bigl), d \biggl) = 1$. From Equation (\ref{p78:ppu78}), we deduce that:
\[
0 \leq \frac{-7 \beta(d)}{8} \leq S3\biggl(\ell+h+d+1+\bigl(\rho(m) \times (-1-p_{d}) \bigl), d \biggl) - \theta_1 + Q2(\ell+L_4(d), d) 
\]
It follows that
\[
z(\ell+L_4(d), d) = 1
\]
i.e. $z(\ell+L_4(d), d) = w\biggl(\ell+h+d+1+\bigl(\rho(m) \times (-1-p_{d}) \bigl), d \biggl)$.  \\

{\it Subcase 2:} $S3\biggl(\ell+h+d+1+\bigl(\rho(m) \times (-1-p_{d}) \bigl), d \biggl) \ < \theta_1$  \\
The hypothesis $S3\biggl(\ell+h+d+1+\bigl(\rho(m) \times (-1-p_{d}) \bigl), d \biggl)  < \theta_1$ implies that
 $w\biggl(\ell+h+d+1+\bigl(\rho(m) \times (-1-p_{d}) \bigl), d \biggl) = 0$. From Remark \ref{r:n2} and  the hypothesis, we deduce that:
\[
S3\biggl(\ell+h+d+1+\bigl(\rho(m) \times (-1-p_{d}) \bigl), d \biggl) - \theta_1  \leq \lambda
\]
From Equation (\ref{p78:ppu78}) we deduce that:
\[
S3\biggl(\ell+h+d+1+\bigl(\rho(m) \times (-1-p_{d}) \bigl), d \biggl) - \theta_1 + Q2(\ell+L_4(d), d) \leq \lambda - \xi(d) < 0
\]
It follows that:
\[
z(\ell+L_4(d), d) = 0 = w\biggl(\ell+h+d+1+\bigl(\rho(m) \times (-1-p_{d}) \bigl), d \biggl).
\]

{\bf Second Case:} $\rho(m)-d-1 \leq i_2 \leq \rho(m)-1$  \\

From induction hypothesis and Remark \ref{rrur:uu5}, it is clear that:
\begin{equation}  \label{eeq:dec09}
\sum_{f=1}^{h} b_f w\biggl(-f+\ell+h+d+1+\bigl(\rho(m) \times (-1-p_{d}) \bigl), d \biggl) = 0 
\end{equation}
Consequently, we have:
\[
w\biggl(\ell+h+d+1+\bigl(\rho(m) \times (-1-p_{d}) \bigl), d \biggl) = 0.
\]
Based on Equation (\ref{p78:ppu78}) and on Equation (\ref{eeq:dec09}), we deduce that:
\[
S3\biggl(\ell+h+d+1+\bigl(\rho(m) \times (-1-p_{d}) \bigl), d \biggl) - \theta_1 + Q2(\ell+L_4(d), d) \leq -\theta_1 - \xi(d)   \ < 0
\]
From Equation (\ref{eeq:n6p29}), we deduce that:
\[
z(\ell+L_4(d), d) = 0.
\]
It follows that:
\[
w\biggl(\ell+h+d+1+\bigl(\rho(m) \times (-1-p_{d}) \bigl), d \biggl) = z(\ell+L_4(d), d).
\]
$\blacksquare$  \\

{\bf Proof of Lemma \ref{ecc:n1} } \\  
The result is deduced from the fact that the
transient phase finishes at time $L_4(d)$ and from the fact that 
the period of the sequence $\{ w(n, d) : n \geq 0 \}$ is $L_0(d)$.
$\blacksquare$  \\

{\bf Proof of Lemma \ref{ch29:eq4} } \\  
Based on Equation (\ref{eeq:n11}), on Equation (\ref{eeq:n23}) and from the following hypothesis:
\[
 d \ < \ \beta(m, \alpha_e),
\]
we easily deduce that:
\[
c(f, d) \ = \ 0 \ ; \  {\rm for \ } f \ {\rm such \ that \ } h-\beta(m, \alpha_e)+d+1 \leq f \leq h.
\]

From one of the following initial configurations $y(0)y(1) \dots y(h-1)$ or  $\widetilde{z}(0,d) \widetilde{z}(1,d) \dots \widetilde{z}(h-1,d)$,
the neuronal recurrence equation $\{ z(n, d) \ : \ n \geq 0 \}$ converges to the same basin of attraction because

\[
\widetilde{z}(\ell, d) = y(\ell) \ {\rm for} \ \ell \ {\rm such \ that \ } \ \beta(m, \alpha_e)-d \leq \ell \leq h-1
\]
and 
\[
c(f, d) \ = \ 0 \ ; \  {\rm for \ } f \ {\rm such \ that \ } h-\beta(m, \alpha_e)+d+1 \leq f \leq h.
\]
$\blacksquare$  \\

\end{document}